\documentclass[runningheads]{llncs}

\usepackage{eccv}

\usepackage{eccvabbrv}

\usepackage{graphicx}
\usepackage{booktabs}

\usepackage[accsupp]{axessibility}  

\usepackage[export]{adjustbox}
\usepackage{multirow}%
\usepackage{subcaption}
\usepackage{microtype}
\usepackage{longtable}

\newcommand{\boldhline}{\specialrule{0.15em}{0em}{0.1em}}

\usepackage[breaklinks,colorlinks,citecolor=eccvblue]{hyperref}

\usepackage[capitalize,nameinlink,sort&compress]{cleveref}

\usepackage{orcidlink}

\begin{document}

\title{CLEO: Continual Learning of Evolving Ontologies} 

\titlerunning{CLEO: Continual Learning of Evolving Ontologies}


\author{Shishir Muralidhara\inst{1} \and
	Saqib Bukhari\inst{2} \and
	Georg Schneider\inst{2} \and
	Didier Stricker\inst{1,3} \and
	Ren\'e Schuster\inst{1,3}}

\authorrunning{S. Muralidhara et al.}


\institute{DFKI -- German Research Center for Artificial Intelligence \and
		ZF Friedrichshafen AG \and
		RPTU – University of Kaiserslautern-Landau \\ 
		\email{\{firstname.lastname\}@\{dfki.de, zf.com\}}}

\maketitle

\begin{abstract}
  Continual learning (CL) addresses the problem of catastrophic forgetting in neural networks, which occurs when a trained model tends to overwrite previously learned information, when presented with a new task.
  CL aims to instill the lifelong learning characteristic of humans in intelligent systems, making them capable of learning continuously while retaining what was already learned. 
  Current CL problems involve either learning new domains (domain-incremental) or new and previously unseen classes (class-incremental). 
  However, general learning processes are not just limited to learning information, but also refinement of existing information.
  In this paper, we define CLEO -- Continual Learning of Evolving Ontologies, as a new incremental learning setting under CL to tackle evolving classes. 
  CLEO is motivated by the need for intelligent systems to adapt to real-world ontologies that change over time, such as those in autonomous driving.
  We use Cityscapes, PASCAL VOC, and Mapillary Vistas to define the task settings and demonstrate the applicability of CLEO.
  We highlight the shortcomings of existing CIL methods in adapting to CLEO and propose a baseline solution, called Modelling Ontologies (MoOn).
  CLEO is a promising new approach to CL that addresses the challenge of evolving ontologies in real-world applications.
  MoOn surpasses previous CL approaches in the context of CLEO.
  
\end{abstract}

\section{Introduction}
\label{sec:intro}

\begin{figure*}[t]
	\includegraphics[width=\textwidth]{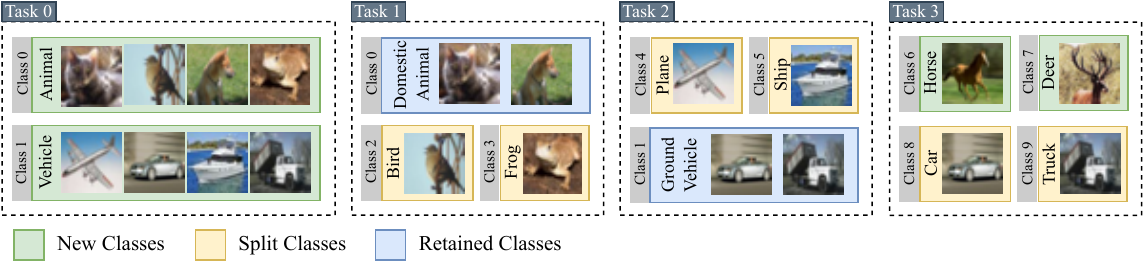}
	\caption{CLEO extends class-incremental learning (CIL) by removing the constraint that incremental sets of classes must be non-overlapping. This allows for more flexible and nuanced learning, as existing classes can be refined into more specific classes. }
\label{fig:cleo_teaser}
\vspace{-0.5em}
\end{figure*}

Advancements in AI have led to the emergence of systems that can match or even outperform human abilities in various domains. However, these systems are often confined to specific datasets and tasks and their efficacy is affected when faced with new tasks or data. This presents a significant hurdle in practical real-world AI applications where the circumstances are constantly evolving. For example, a self-driving car needs to adapt to new traffic regulations, signage, and weather conditions. When a trained model encounters new information or tasks, it experiences a phenomenon known as \textit{catastrophic forgetting} \cite{catastrophic_forgetting}. This issue arises from the rigidity of neural networks, where previously learned knowledge is overwritten when learning something new.
Consequently, the model's performance significantly deteriorates, which can have severe repercussions in scenarios where the model has to acclimate to dynamic environments or critical applications where performance is imperative. Therefore, tackling forgetting is crucial when developing AI systems with lifelong learning.

In contrast, learning in humans is continuous and dynamic through which we continually assimilate new information while preserving and enriching previously learned knowledge. 
The goal of CL is to instill neural networks with this learning characteristic and make them capable of extending when objectives change. 
In incremental learning, which involves learning a sequence of tasks, the term \textit{task} can represent changes either in the distribution of data, referred to as domain-incremental learning (DIL), or changes in the output space, which is known as class-incremental learning (CIL). 
However, learning is not limited only to the acquisition of new knowledge; it also involves refining and deepening our grasp of existing information.
To illustrate, our initial understanding of animals might have been confined to recognizing a broad categorization, like distinguishing between birds and other animals. However, over time, our learning process matures to incorporate finer, more nuanced insights about individual animals.
This highlights the necessity of broadening the scope of CL to accommodate both the introduction of novel classes, and also the enhancement of existing knowledge.
\newline\newline
We propose a novel incremental setting termed CLEO -- Continual Learning of Evolving Ontologies.
CLEO represents a progressive extension of CIL, overcoming the confining assumption that all newly encountered classes are entirely novel and previously unseen.
CLEO acknowledges the dynamic nature of learning, where classes may not only emerge but also evolve over time.
CLEO combines elements of novelty and refinement, paving the way for more resilient AI systems.
The real-world relevance of CLEO is exemplified by the Mapillary Vistas dataset \cite{neuhold2017mapillary}, where an initial set of classes has evolved into more fine-grained and specific classes.
An illustration of CLEO is presented in \cref{fig:cleo_teaser} using the CIFAR-10 dataset \cite{krizhevsky2009learning} as an example. In the initial task, the system learns two broad categories, namely \textit{animals} and \textit{vehicles}. Subsequently, more specific classes, such as \textit{bird, frog, plane, and ship}, are split from the original classes. CLEO presents a challenge where the system must not only learn the evolved classes, which it had previously encountered within a different semantic context, but also retain the remaining information.
In this example, after \textit{bird and frog} were separated from the \textit{animal} class, \textit{cat} and \textit{dog} are left as the remaining \textit{domestic animal} class.
As in CIL, CLEO also covers the case that entirely new classes are learned, like \eg the classes \textit{horse} and \textit{deer} in the last task of \cref{fig:cleo_teaser}. 
CLEO embodies the fundamental idea that AI systems must possess the ability to adapt to dynamic shifts in which classes emerge, evolve, and split over time.
By integrating this adaptability into AI systems, we enhance their robustness in real-world applications.
This paper makes several significant contributions:
\begin{itemize}
\item \textbf{Formalizing Continual Learning of Evolving Ontologies:} A primary contribution is the introduction and formalization of the concept of learning evolving ontologies, from a continual learning perspective.

\item \textbf{Defining Task Settings:} To provide a comprehensive understanding of CLEO, we define several task settings for semantic segmentation, using three diverse datasets: Cityscapes \cite{cordts2016cityscapes}, PASCAL VOC \cite{everingham2010pascal}, and Mapillary Vistas \cite{neuhold2017mapillary}.

\item \textbf{Evaluation Metrics for CLEO:} By identifying class groups and measuring their performance, we define a more nuanced evaluation of how effectively the system learns refined classes and retains information.    

\item \textbf{Baseline Solution -- MoOn:} To operationalize the concept for semantic segmentation, we propose a baseline solution called "Modelling Ontologies" (MoOn), to handle the challenges inherent in CLEO. 

\end{itemize}


\section{Related Work}

\subsection{Continual Learning}
Continual learning (CL), also known as lifelong learning, and its sub-field of incremental learning, are critical areas of machine learning that address a fundamental limitation of traditional machine learning models: Their rigidity in learning new tasks.
CL is a learning paradigm that aims to train models that can learn new tasks and information without forgetting what was previously learned. This contrasts traditional machine learning, where models are typically trained on a single dataset and then remain fixed.
Continual learning presents a challenge of balancing between learning new information and retaining the old \cite{stability-plasticity}.
Overemphasizing learning new information leads to forgetting the old while focusing too much on retaining old information hinders learning the new task.
CL approaches can be grouped into three categories: Architecture-based, replay-based methods, and regularization-based methods.

Architecture-based approaches modify the architecture of neural networks to accommodate new tasks without overwriting previous information. 
These include methods such as parameter isolation \cite{piggyback, packnet, pathnet}, dynamically growing networks \cite{progressiveNN, ExpertGate, DEN}, or incorporation of additional components such as memory modules \cite{progressivebanks, MBPA}. 
These methods can face scalability challenges as network architecture grows, increasing memory and computational demands.

Replay-based approaches are of two types: Experience replay or rehearsal, involves storing a few instances from the previous task \cite{rainbow_memory, ringBuffer, GEM}, which are used during training on a new task. 
Generative replay substitutes the memory buffer with a generative model to sample previous task instances when learning on a new task \cite{fearnet, van2020brain, DGR}.
Challenges include privacy, data availability, memory and computational constraints in training generative models.

Regularization-based approaches \cite{EWC, SI, LWF} strike a balance between stability and plasticity when learning new tasks by introducing a regularization term into the loss function. One of the approaches to regularization is to determine the importance of parameters to the previous tasks and penalize overwriting them. 
Distillation is another approach which preserves previously learned knowledge by distilling information from the old model to the new.

\subsection{Continual Semantic Segmentation}

Continual semantic segmentation (CSS) \cite{SSUL, selfTrainingCIL, continualAttnFusion, contrastiveDistillation} is a specialized domain within CL that focuses on the task of semantic segmentation. Incremental Learning Techniques \cite{ILT} (ILT) is a regularization-based method that distils knowledge from previously trained task models and additionally freezes the encoder for the current task, which acts as an additional regularization constraint.
Modeling the Background \cite{MiB} (MiB) addresses the problem of background shift in incremental segmentation.  It proposes a novel distillation loss that accounts for the changing background between tasks.
AWT \cite{AWT} addresses the background shift in classifier initialization, and transfers the most relevant weights from the previous classifiers to the new classifier.
In our baseline solution for CLEO, we will build on the ideas of MiB and extend it to model the semantic shift for all classes.  
PLOP \cite{PLOP} argues that treating the background class as a single class as done in MiB, which includes both previously encountered and novel classes can lead to catastrophic forgetting of the old classes as it learns new classes. PLOP addresses this problem by pseudo-labeling old classes in the background. 
EWF \cite{EWF} combines the weights of the old and new model trained with regularization to give a more balanced model.
SDR \cite{SDR} preserves latent representations of old classes through prototype matching and frees space for new classes through feature sparsity and contrastive learning. The decoder preserves previous knowledge via output-level distillation.
RCIL \cite{RCIL} uses a representation compensation module to decouple the representation learning of old and new classes. This helps to prevent the model from forgetting old classes when learning new classes.
REMINDER \cite{reminder} proposes a class-similarity weighted distillation that distills knowledge from a previous model on old classes that are similar to the new classes. This reduces forgetting of old classes and improves learning of new classes.



\subsection{Evolving Ontologies}

An emerging area within incremental learning focuses on learning evolving classes, which involves the refinement of existing knowledge, where a broad initial grouping of classes evolves over time into more specific and nuanced classes. In contrast, in CIL, each task consists of a non-overlapping subsets of novel classes. 
\newline\newline
Hierarchical Label Expansion \cite{HLE} (HLE) for classification involves a model initially learning coarse classes, which then evolve into more fine-grained categories during each task across varying hierarchical levels.
The expansion is restrictive, where the entire subset of classes at the current level expands in the next level during each task. 
CLEO differs from HLE in the following aspects:
\begin{itemize}
	
	\item CLEO allows a more flexible evolution 
	at any point within the sequence of tasks, in which certain classes can remain (partially) unexpanded or expand into sub-groups.
	
	\item The baseline solution for HLE is a rehearsal-based method that uses pseudo-labeling from previous classifiers -- one for each level of the hierarchy -- for determining the importance of data.
	We present a regularization-based solution that uses knowledge distillation.
	
	\item HLE considers image classification as application, while CLEO additionally deals with the more challenging problem of semantic segmentation.
	
\end{itemize}


Learning with Evolving Class Ontologies \cite{LECO} (LECO) has been the first attempt to address evolving ontologies in semantic segmentation. However, despite its intended framing as a continual learning problem, LECO deviates from the defining principles of continual learning. We address these shortcomings and define this problem strictly within the bounds of continual learning. The primary distinctions between our framework and LECO are outlined below:

\begin{itemize}
	
	\item LECO requires and stores all historical data, a constraint that contradicts the definition of CL \cite{clRobotics}, that all data is not available at once.
	
	\item LECO primarily revolves around the dilemma of whether to label new data or relabel the old data with the current label space. In contrast, CLEO confines itself to utilizing the input data and annotations within the dataset.
	
	\item Due to the availability of historical data, LECO employs naive strategies such as fine-tuning, freezing, or joint training. We highlight the limitations of existing CIL methods and propose a baseline specifically for CLEO.
	
	\item CLEO explicitly addresses catastrophic forgetting, which is one of the core problems of continual learning. Since LECO retrains entirely using all the data, the problem of catastrophic forgetting does not even arise.
\end{itemize}

\section{CLEO}
Deep learning has drawn inspiration from biological systems at various levels, from the way neurons form the fundamental building blocks of artificial neural networks to learning paradigms such as curriculum learning and reinforcement learning.
Continual learning is another learning paradigm that aims to instill lifelong learning in intelligent systems.
As discussed previously, current problems in continual learning involve adaptation to different domains, such as handling adverse conditions in autonomous driving or learning new classes over time.
A common characteristic in both these scenarios is that the task always represents new and previously unseen data.
However, it is our contention that the scope of continual learning should not be limited solely to the discovery of novel classes. It should also include the enhancement and refinement of existing knowledge. This aligns closely with the way humans continually adapt and refine knowledge through neuroplasticity.

\begin{figure}[t]
	\centering
	
	 \begin{subfigure}{0.32\linewidth}
		     \includegraphics[width=\linewidth]{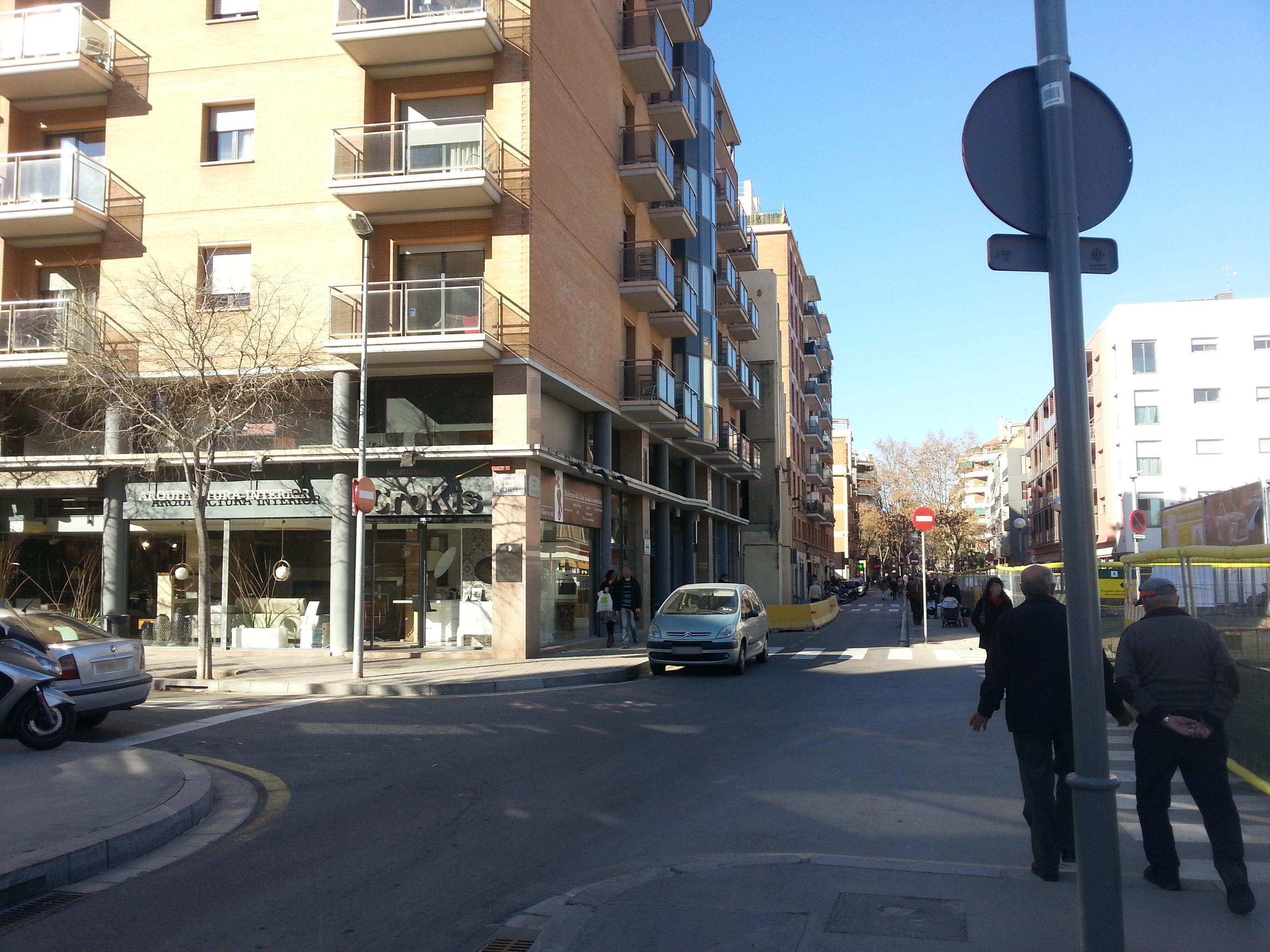}
		 \end{subfigure}\hspace*{\fill}
	\begin{subfigure}{0.32\linewidth}
		\includegraphics[width=\linewidth]{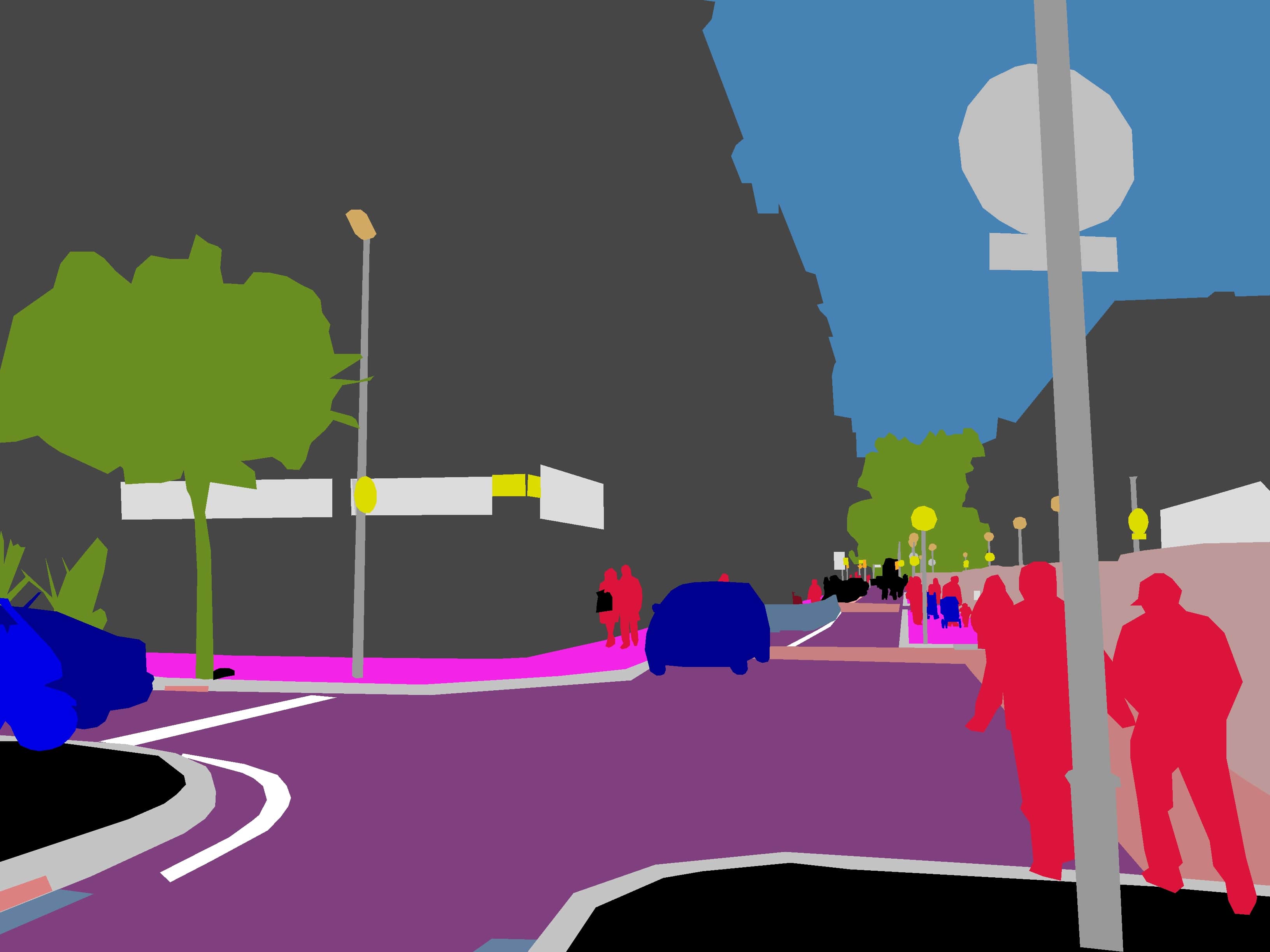}
	\end{subfigure}\hspace*{\fill}
	\begin{subfigure}{0.32\linewidth}
		\includegraphics[width=\linewidth]{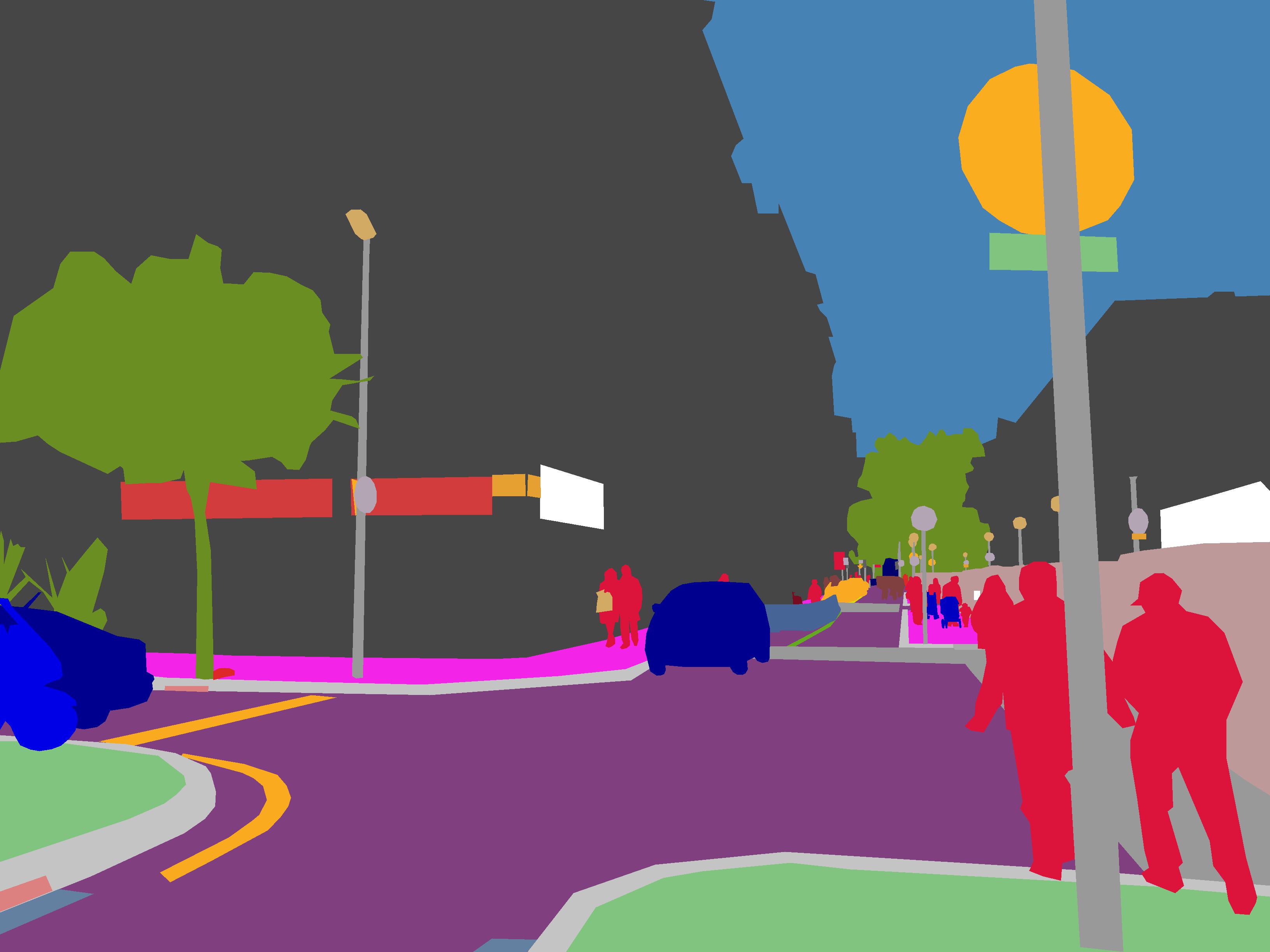}
	\end{subfigure}
	
	\caption{The Mapillary Vistas dataset \cite{neuhold2017mapillary} provides a real-world example for an evolving ontology. This figure compares ground truth annotations for the same image in two dataset versions. It illustrates the introduction of new classes, such as \textit{traffic island} and the finer categorization of existing classes such as \textit{billboard}. }

\label{fig:evolving_ontologies}
\end{figure}

In light of this perspective, we introduce CLEO -- Continual Learning of Evolving Ontologies. CLEO is centered around the idea of removing the constraints of class-incremental learning and allows classes that have a hierarchical relation (partially) overlap between tasks, thereby facilitating the refinement of already learned information.
CLEO is a real and open-world problem and this is exemplified in the Mapillary Vistas dataset \cite{neuhold2017mapillary}. The initial version of the dataset defined 66 classes, which later evolved into 124 classes in the latest version. It is noteworthy to observe the transformation within class categorization. On one hand, classes that were previously categorized under \textit{unlabeled}, such as \textit{traffic island} emerge as novel classes. 
Conversely, several pre-existing classes undergo refinement, leading to the creation of new, more specific categories.
In \cref{fig:evolving_ontologies}, we can observe a tangible example of this evolution, where the previous \textit{billboard} class has evolved into more specialized classes, including signage classes for advertisements, information, stores, and differentiating between the front and back of the signs. 
In critical applications like autonomous driving, it becomes imperative for AI systems to swiftly adapt to these nuanced changes, perpetually learn, and enhance scene understanding precision.
\newline\newline
Let $T = \{0, 1, 2,...,n\}$ denote a sequence of $n+1$ tasks.
Each task is associated with a specific subset of classes denoted as $C_t$.
The union of all classes across tasks is $C =  \bigcup_{t=0}^{n} C_t$, representing the entire set of classes encountered during the learning process. At each task $t$, a subset of classes $C_t$ is introduced incrementally similar to CIL.
$C_{0:t} = C_0 \cup ... \cup C_{t-1} \cup C_t$ is the set of all classes seen until step $t$.
CLEO acknowledges the possibility of class splitting, where an existing class $c_{t-1}$ from $C_{0:t-1}$ can evolve into multiple new classes in $C_t$.
Formally this can be expressed as, $C_t \subset C$, $C_i \cap C_j = \phi$ for $i \neq j$ and the class splitting is expressed as $c_{t-1} \in C_{0:t-1} \rightarrow \{c_{k+1}, c_{k+2},...,c_{k+m}\} \subseteq C_t$.
Note that the splitting of a class is not necessarily exhaustive, \ie $c_{t-1} \in C_{0:t-1}$ may persist.
An example for such an evolution of classes is given in \cref{fig:cleo_teaser}.
The learning objective of a task remains consistent with CIL: Train a model to recognize and differentiate the current subset of classes $C_t$, while ensuring that the model retains knowledge from the previously encountered classes in $C_{0:t-1}$. As in CIL, the model is not provided access to data from previous tasks. 

\section{MoOn: Modelling Ontologies for CLEO}
Class-incremental learning involves the process of training a model to incrementally learn new classes. At each increment, the model is presented with images and the corresponding ground truth which contains annotations for the new classes being learned.
In class-incremental semantic segmentation (CISS), at each step, all unlabeled areas are marked as \textit{background}. Thus in CISS, the phenomenon of \textit{semantic background shift} occurs, because in each task, the meaning of the \textit{background} class ($bg$) changes.
Distillation loss is a technique of transferring knowledge from an old model $f_{t-1}$ to a new model $f_t$, and this is achieved by comparing the old model's prediction to the new model's predictions. The distillation loss enforces the new model's predictions to resemble those of the old model by penalizing any deviations and this helps mitigate forgetting.
Consider an image $x$ with $N$ pixels and the corresponding ground truth $y$ at the current task $t$ with the class set $C_t$.
The standard distillation loss in \cite{MiB} is defined as follows:
\begin{equation}
	\ell_{kd}^t (x,y) = - \frac{1}{N} \sum_{i \in x} \sum_{c\in C_{0:t-1}} p_x^{t-1}(i,c) \log \hat{p}_x^t(i,c)
\end{equation}
where $p_x^{t-1}(i,c)$ is the probability for class $c$ in pixel $i$ given by the model $f_{t-1}$ and $\hat{p}_x^{t}(i,c)$ is for class $c$ in pixel $i$ given by the model $f_{t}$ but re-normalized across classes $C_{0:t-1}$: 
\begin{equation}
	\hat{p}_x^t(i,c) = \begin{cases}
		0 & \text{if } c \in C_t \setminus\{bg\} \\
		\dfrac{p_x^{t}(i,c)}{\sum_{k \in C_{0:t-1}}p_x^{t}(i,k)} & \text{if } c \in C_{0:t-1}
	\end{cases}
	\label{eq:std_dloss}
\end{equation}
Standard distillation ignores the dynamic nature of the background in incremental semantic segmentation. The \textit{background} class in segmentation is a catch-all class that includes all previously seen classes and potential future classes. Furthermore, the background class is present in all tasks and changes as new classes are added, leading to background shift. A class $c \in C_t$ added during the current task was previously part of the \textit{background} in $C_{t-1}$. For a pixel representing a new class at the current task $t$, the old model $f_{t-1}$ would predict the \textit{background} class, and if standard distillation loss was applied naively, it would penalize the new model $f_t$  for predicting the correct class.
\begin{figure*}[t] 
	\centering
	\includegraphics[width=\textwidth]{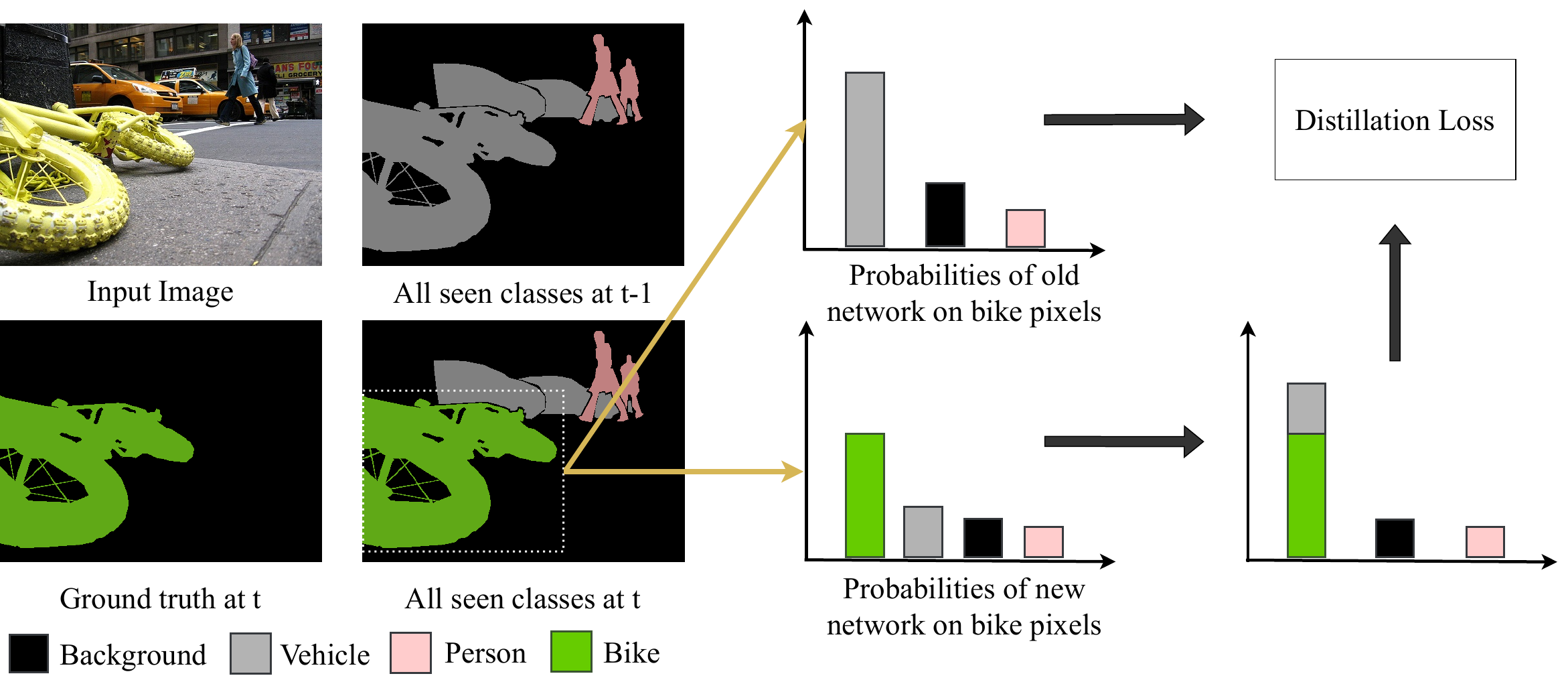}
	\caption{Modelling Ontologies (MoOn) generalizes the idea of MiB \cite{MiB} for all classes. During distillation, forgetting is mitigated, by grouping class-logits according to the evolution of the class ontology. In this example, the teacher network of the previous task predicts the \textit{vehicle} class, while the model tries to learn \textit{bicycle}. To avoid penalizing this, the predictions for \textit{vehicle} and \textit{bicycle} are combined before the distillation loss.}
	\label{fig:MoOn}
\end{figure*}
Modeling the Background (MiB) \cite{MiB} proposes a novel distillation loss that explicitly addresses the background shift in incremental segmentation. MiB does not directly compare the prediction of the new model $f_t$ with the corresponding prediction of the old model $f_{t-1}$.
Instead, the background prediction by $f_{t-1}$ is compared with the $f_t$ prediction of either a new class \emph{or} the background. \cref{eq:mib_dloss} modifies \cref{eq:std_dloss} by additionally considering the whole label space instead of limiting it to $C_{0:t-1}$.
MiB acknowledges that pixels that were thought to be part of the background by the old model might actually be the new classes the current model is trying to learn, thus avoiding penalizing the probabilities of new classes being learned.
\begin{equation}
	\hat{p}_x^t(i,c) = \begin{cases}
		p_x^{t}(i,c) & \text{if } c \neq bg \\
		\sum_{k \in C_t}p_x^{t}(i,k) & \text{if } c = bg
	\end{cases}
	\label{eq:mib_dloss}
\end{equation}
MiB is designed for CIL, where the new classes being learned at the current step, were previously unseen and are always a part of the background class. Therefore, it would suffice to take into consideration the background class, and hence modeling the background. Whereas CLEO is designed for learning of evolving classes and
the new classes being learned may have been previously unseen and part of the background or been part of a broader class. As seen in \cref{fig:MoOn}, the class \textit{bike} being learned at the current step may have been previously learned as a \textit{vehicle} class grouped with other classes like \textit{truck} and \textit{bus}. Therefore we need to take into consideration that within CLEO, new classes may not always have been a part of the \textit{background} class previously.
\newline\newline
To this end, we propose Modelling Ontologies (MoOn) as a baseline solution for CLEO, which can be seen as a generalized version of MiB that accounts for the fact that new classes may emerge from any existing class, and not just the background class.
Similar to how MiB considered the possibility of the new class pixel belonging to the background class previously, MoOn additionally includes the possibility of the pixel belonging to an existing class. In MiB, the old model $f_{t-1}$ may predict the background class for the new class being learned, whereas the old model in MoOn may predict an existing class $c \in C_{0:t-1}$. Therefore, the correct distillation would be to distill the knowledge from the old model $f_{t-1}$ for the existing class $c \in C_{0:t-1}$ that is being split into new classes in $C_t$.  In the illustrated scenario depicted in \cref{fig:MoOn}, distillation would involve transferring knowledge from the old model regarding the \textit{vehicle} class, while training the new model to predict the \textit{bike} class.
To achieve this, at each step~$t$, MoOn uses a set $S_t$ of class maps $(c_{t-1} \rightarrow \{c_{k+1}, c_{k+2},...,c_{k+m}\} = C_{new})$ to identify the class $c_{t-1} \in C_{0:t-1}$ from which the current classes in $C_{t}$ are split. If the new class $c_t$ is completely new, 
then the superclass $c_{t-1}$ in the class mapping would be the \textit{background} class $bg$.
We define $C_{e,t}$ as the set of all evolving classes $c_{t-1}$ in $S_t$.
The modified distillation loss for learning evolving classes in CLEO is given as:
\begin{equation}
	\centering
	\hat{p}_x^t(i,c) = 
	\begin{cases}
		p_x^{t}(i,c) & \text{if } c \notin C_{e,t} \vspace{0.5em} \\
		\begin{aligned} p_x^{t}(i,c_{t-1}) + \sum_{c_n \in C_{new}} p_x^{t}(i,c_n)\end{aligned} & \text{if } c \in C_{e,t}
	\end{cases}
	\label{eq:moon_dloss}
\end{equation}


\subsection{Datasets}
Towards the introduction of CLEO for semantic segmentation, we consider the following three diverse datasets:

\begin{itemize}
	
	\item \textbf{Cityscapes:}  Cityscapes \cite{cordts2016cityscapes} is a large-scale dataset of urban street scenes captured in 50 different cities under daytime and good weather conditions. The dataset provides annotations for 30 classes out of which 19 are used for evaluation and grouped into 7 categories. An overview of the hierarchy of the 19 semantic classes grouped by their parent categories is presented in the supplementary.
	
	\item \textbf{PASCAL VOC:} PASCAL VOC 2012 \cite{everingham2010pascal} is part of the Visual Object Classes Challenge, which ran from 2005 to 2012. PASCAL VOC is a more general dataset like COCO \cite{lin2014coco} consisting of 20 categories, including vehicles, household objects, animals, and persons. The categorization by \cite{VOC_Hierarchy} of these classes along with intermediate groupings is shown in the supplementary. 
	
	\item \textbf{Mapillary Vistas:} Mapillary Vistas \cite{neuhold2017mapillary} is a large-scale street-level image dataset for understanding street scenes around the world. It covers a wider range of geographic locations and diverse weather and illumination conditions. The initial version of the dataset provided annotations for 66 classes, which later nearly doubled into 124 classes in the current version v2.0.

\end{itemize}

\subsection{Baselines and Implementation}

We evaluate our method’s performance against standard CL baselines, namely \textit{fine-tuning} and \textit{joint training}. Fine-tuning involves using the previous step model as the starting point when learning on the current task. This is similar to transfer learning and even results in positive forward transfer. However, since there is no explicit attempt to prevent catastrophic forgetting, it involves the highest amount of forgetting and forms the lower baseline. The joint training model involves learning all of the tasks together in a single step, and as there is no incremental learning there is zero forgetting and this forms the theoretical upper baseline.
Since in CLEO a redefinition of classes may occur because of the evolving ontology, we define the set of classes for the \textit{joint training} as the final set of classes after the last task.
Additionally, we compare against the state-of-the-art (SOTA) methods such as MiB \cite{MiB}, PLOP \cite{PLOP} and RCIL \cite{RCIL}.
\newline\newline
For MiB, and our proposed approach, we adopt the framework proposed by MiB \cite{MiB}. In the case of PLOP and RCIL, which involve modifications specific to their approaches, we adopt the RCIL framework \cite{RCIL}. All these approaches use the DeepLabV3 \cite{DeepLabV3} architecture as segmentation network, utilizing a ResNet-101 \cite{ResNet} backbone with an output stride of 16. The backbone’s initialization uses a pre-trained model from ImageNet \cite{ImageNet}. The training methodology utilizes SGD, with a learning rate of 0.01 for the initial learning step and 0.001 for subsequent steps. Across all learning steps, we train the model using a batch size of 24, for 50 epochs on Cityscapes and for 30 epochs on PASCAL and Mapillary Vistas.

\subsection{Evaluation Metrics} \label{sec:exp:metrics}

Mean Intersection-over-Union (mIoU) is a commonly used metric for evaluating semantic segmentation models. It measures the average overlap between the predicted segmentation and the ground truth segmentation for all classes. To calculate the mIoU, the Intersection-over-Union (IoU) is first calculated for each class, and then the IoU scores for all classes are averaged. The performance of CL models are often evaluated by measuring the amount of information forgotten by comparing the model's performance on old tasks against the joint training baseline.
Within the context of CLEO, while learning evolving classes, forgetting can occur at multiple levels. We identify three types of class groups and analyze the mIoU results for each of them:
\begin{itemize}
	
	\item \textbf{Unsplit Classes:} These classes are learned in the first task and are not further split. Their performance is expected to be largely unaffected, as they have not been split nor learned incrementally.
	
	\item \textbf{Split Classes:}  These classes have been split either from the background or a previously learned superclass. The mIoU results for these split classes will provide insight into the model's ability to learn new classes.
	
	\item \textbf{Retained Classes:} These are classes remaining in the parent class after it has been split partially to yield new classes. Analyzing the mIoU scores for retained classes sheds light on how well the model retains knowledge.
\end{itemize}


\subsection{Experimental Settings and Results}

We introduce seven initial experimental settings, that are designed for the previously introduced datasets and resemble the complexity and diversity of CLEO beyond classical CIL.
For completeness and better understanding, we present the detailed semantic evolution of all settings by listing $C_t$ for every task in the supplementary, including the class groups as introduced in \cref{sec:exp:metrics}. 
\section{Experiments}

\begin{table*}[t]
	\centering
	\caption{Results for our two CLEO settings on Cityscapes \cite{cordts2016cityscapes} after learning all tasks.}
	\begin{tabular}{c||cccc||cccc}
		\boldhline
		
		\multirow{2}{*}{\textbf{Method}} & \multicolumn{4}{c||}{\textbf{CS-Ex1}} & \multicolumn{4}{c}{\textbf{CS-Ex2}} \\ 
		& \textit{Unsplit} & \textit{Split} & \textit{Retained} & \textbf{All} & \textit{Unsplit} & \textit{Split} & \textit{Retained} & \textbf{All} \\ \boldhline
		
		Fine-Tuning & 00.00 & 00.00 & 00.00 & 00.00 & 00.00 & 14.46 & 00.00 & 08.68 \\
		\hline
		Joint Training & 90.26 & 61.15 & 48.04 & 55.62 & 90.26 & 61.15 & 48.04 & 55.62  \\
		
		\boldhline
		
		MiB \cite{MiB}& 90.24 & 03.66 & 07.99 & 09.11 & 90.23 & 18.25 & 11.39 & 18.88 \\ \hline
		
		PLOP \cite{PLOP}& 87.58 & 34.36 & 13.02 & 28.91 & 86.01 & 31.91 & 15.51 & 28.10  \\ \hline
		
		RCIL \cite{RCIL} & 89.82 & 00.00 & 07.97 & 06.88 & 89.00 & 00.00 & 07.70 & 06.76 \\ \hline
		
		MoOn (Ours)& 90.21 & 39.50 & 37.00 & \textbf{39.31} & 90.23  & 42.28 & 35.78 & \textbf{40.62} \\ \boldhline
		
	\end{tabular}
	
	\label{tab:cityscapes}
\end{table*}

\subsubsection{Cityscapes}

\begin{table*}[t]
	\centering
	\caption{Task-wise results for CS-Ex2 after learning the final task.}
	\label{tab:cs_ex2_task_results}
	\begin{tabular}{c || ccccccc ||c}
		\boldhline
		\textbf{Method} & \textbf{ Task 0} & \textbf{Task 1} & \textbf{Task 2} & \textbf{Task 3} & \textbf{Task 4} & \textbf{Task 5} & \textbf{Task 6} & \textbf{All}\\ \boldhline
		Fine-Tuning & 00.00 & 00.00 & 00.00 & 00.00 & 00.00 & 00.00 & 34.70 & 08.68  \\ \hline
		Joint Training & 47.50 & 97.57 & 63.09 & 30.20 & 88.31 & 64.31 & 57.93 & 55.62 \\ \boldhline
		MiB \cite{MiB} & 19.82 & 02.81 & 09.26 & 00.19 & 74.00 & 45.62 & 15.54 & 18.88 \\ \hline
		PLOP \cite{PLOP} & 22.38 & 95.08 & 41.59 & 00.00 & 84.81 & 53.84 & 13.21 & 28.10 \\ \hline
		RCIL \cite{RCIL} & 16.90 & 00.00 & 00.00 & 00.00 & 00.00 & 00.00 & 00.00 & 06.76 \\ \hline
		MoOn (Ours) & 38.11 & 96.89 & 56.26 & 09.19 & 88.16 & 60.97 & 26.10 & \textbf{40.62} \\ \boldhline
	\end{tabular}
\end{table*}

From the Cityscapes class hierarchy, we derive two experimental settings.
Both settings start in task 0 by learning the seven parent classes which together consist of the 19 grouped classes. The settings involve the incremental splitting of individual classes from parent classes. We do not always completely split a parent class, thus emphasizing not only learning the split class but also retaining the remaining information on the superclass.

\begin{itemize}
\item \textbf{CS-Ex1:} Each step involves splitting out a single class from each parent class. Since there is an unequal number of classes in the parent classes, we repeat this process until we have 19 individual classes, resulting in six tasks.


\item \textbf{CS-Ex2:} At each step, we split all but one of the classes from a single parent class to form individual classes. This results in 19 classes in six evolutionary increments (seven tasks in total) for six parent classes excluding \textit{sky}.
\end{itemize}

The results from these settings are presented in \cref{tab:cityscapes} and the task-wise results of \textit{CS-Ex2} in \cref{tab:cs_ex2_task_results}.   
We can observe the shortcomings of existing CIL methods in handling evolving ontologies, whereas our baseline solution, MoOn, outperforms these methods.
All methods achieve similar results on the unsplit classes, which is \textit{sky} in both settings. However, when learning split classes, only PLOP comes close to our approach. Notably, MoOn surpasses PLOP by a significant margin in preserving information within the existing class.
In the second task setting, the last step involves learning five vehicle classes, which are well-represented in the dataset. FT and MIB methods benefit from this, whereas PLOP and RCIL achieve similar results across both settings. 
We provide task-wise results of \textit{CS-Ex1} and visualizations for MoOn in the supplementary.

\begin{table*}[t]
	\centering
	\caption{Results for our three CLEO settings on PASCAL \cite{everingham2010pascal} after learning all tasks.}
	\begin{adjustbox}{width=\textwidth}
		\begin{tabular}{c||cccc||cccc||cccc}
			\boldhline
			
			\multirow{2}{*}{\textbf{Method}} & \multicolumn{4}{c||}{\textbf{VOC-Ex1}} & \multicolumn{4}{c||}{\textbf{VOC-Ex2}} & \multicolumn{4}{c}{\textbf{VOC-Ex3}} \\ 
			& \textit{Unsplit} & \textit{Split} & \textit{Retained} & \textbf{All} & \textit{Unsplit} & \textit{Split} & \textit{Retained} & \textbf{All} & \textit{Unsplit} & \textit{Split} & \textit{Retained} & \textbf{All} \\ \boldhline
			
			Fine-Tuning & 37.77 & 36.31 & 00.00 & 29.68 & 37.74 & 05.38 & 00.00 & 07.81 & 36.91 & 25.68 & 00.00 & 23.08 \\ 
			\hline
			
			Joint Training & 90.19 & 72.53 & 84.12 & 76.91 & 90.48 & 73.78 & 89.14 & 77.75 & 89.84 & 75.27 & 87.36 & 78.38 \\
			
			\boldhline
			
			MiB \cite{MiB} & 88.91 & 37.28 & 54.99 & 47.05 & 86.99 & 11.28 & 37.87 & 22.84 & 87.01 & 35.75 & 33.28 & 40.28 \\ \hline
			
			PLOP \cite{PLOP} & 86.85 & 38.67 & 43.56 & 45.62 & 85.77 & 16.18 & 03.67 & 21.26 & 83.41 & 36.74 & 06.61 & 36.88 \\ \hline
			
			RCIL \cite{RCIL} & 88.08 & 00.84 & 41.11 & 19.29 & 85.29 & 01.43 & 23.85 & 13.18 & 87.45 & 01.99 & 18.63 & 12.51 \\ \hline
			
			MoOn (Ours) & 88.95 & 55.98 & 68.84 & \textbf{62.52} & 84.45 & 22.08 & 25.42 & \textbf{28.82} & 83.33 & 47.51 & 42.36 & \textbf{50.19} \\ \boldhline
			
		\end{tabular}
	\end{adjustbox}
	\label{tab:pascal}
\end{table*}

\subsubsection{PASCAL VOC}

From the PASCAL class hierarchy, we derive three experimental settings that represent the challenges of CLEO. In the first step (task 0), the model learns all the 20 classes grouped into the four parent classes namely \textit{animals}, \textit{household}, \textit{person}, and \textit{vehicle}. The \textit{person} class does not contain any subclasses and therefore does not undergo splitting.

\begin{itemize}
\item \textbf{VOC-Ex1:} This setting includes splitting into intermediate sub-groups. Task 1 includes learning the sub-groups \textit{Farmyard, furniture,} and \textit{2-wheeler} which in turn contain individual classes. In the next and final step, these sub-groups are split into their individual classes. 

\item \textbf{VOC-Ex2:} One class is split sequentially from each parent class resulting in five increments. At the end of the learning, we retain the individual classes \textit{cat}, \textit{bottle}, and the subgroup \textit{4-wheeler} in their respective parent classes.


\item \textbf{VOC-Ex3:} This setting involves splitting all but one of the classes from each parent class sequentially, resulting in four tasks, with three evolutionary steps for
the three parent classes excluding \textit{person} which remains unsplit. 
\end{itemize}

\begin{figure*}[t]
	\centering
	\setlength{\tabcolsep}{1pt}
	
	\begin{adjustbox}{width=\textwidth}
		
		\begin{tabular}{cccccccc}
			\small
			\textbf{Image} & \textbf{GT} & \textbf{FT} & \textbf{JT} & \textbf{MiB} \cite{MiB} & \textbf{PLOP} \cite{PLOP} & \textbf{RCIL} \cite{RCIL} & \textbf{MoOn} \\

			\includegraphics[width=.12\linewidth]{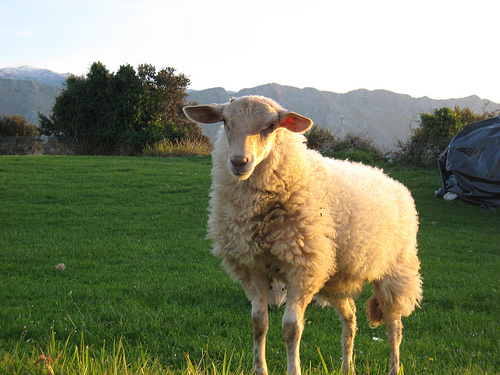} &
			\includegraphics[width=.12\linewidth]{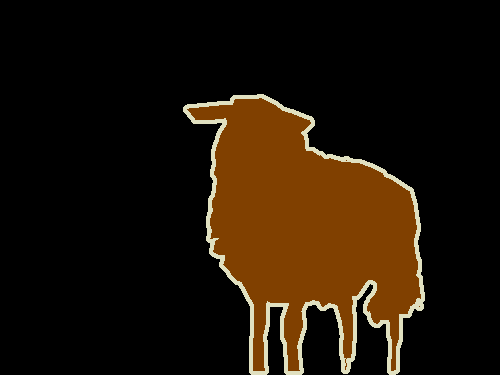}& 
			\includegraphics[width=.12\linewidth]{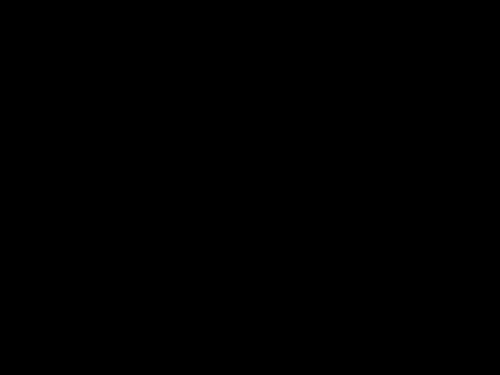}& 
			\includegraphics[width=.12\linewidth]{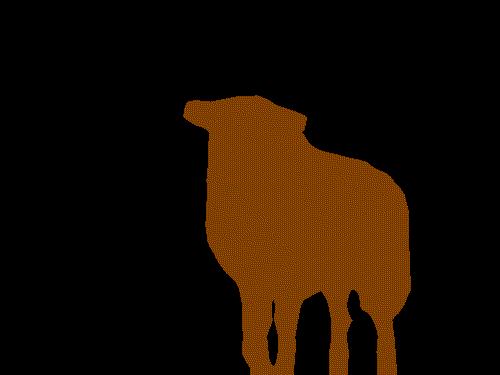} & 
			\includegraphics[width=.12\linewidth]{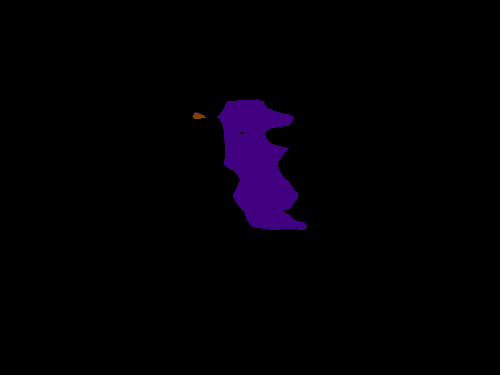} & 
			\includegraphics[width=.12\linewidth]{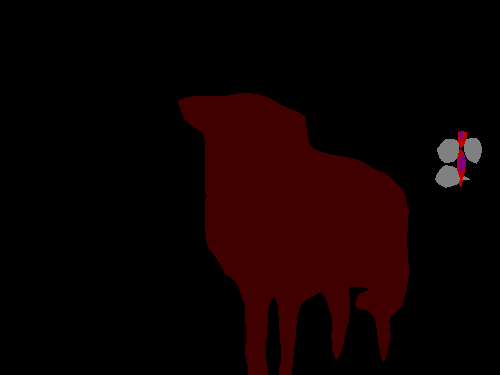}& 
			\includegraphics[width=.12\linewidth]{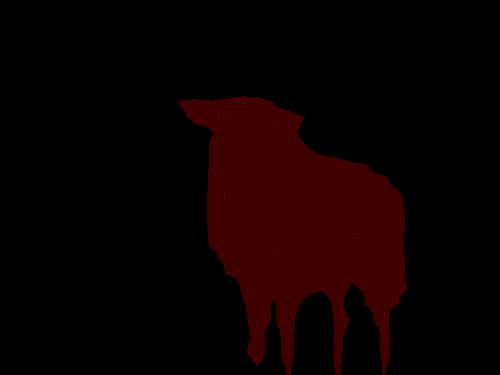} & 
			\includegraphics[width=.12\linewidth]{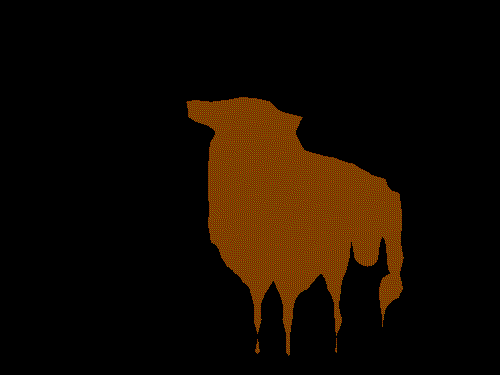}\\[-0.4ex]
			
			\includegraphics[width=.12\linewidth]{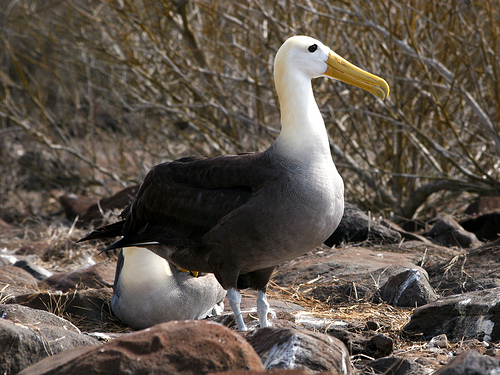} &
			\includegraphics[width=.12\linewidth]{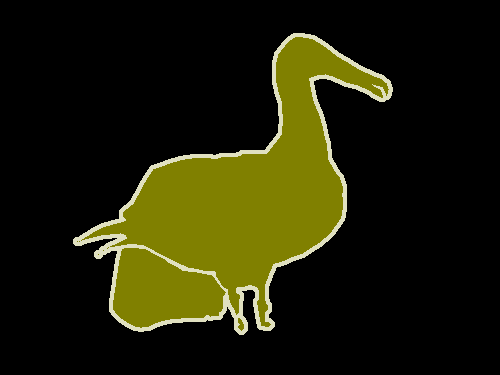}& 
			\includegraphics[width=.12\linewidth]{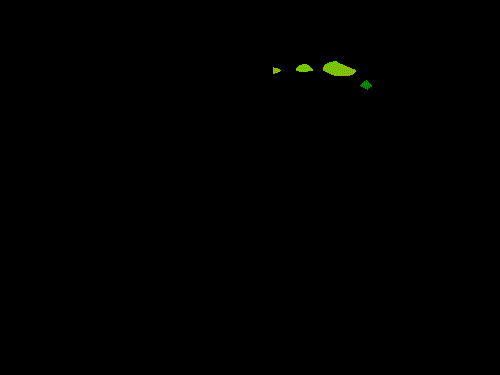}& 
			\includegraphics[width=.12\linewidth]{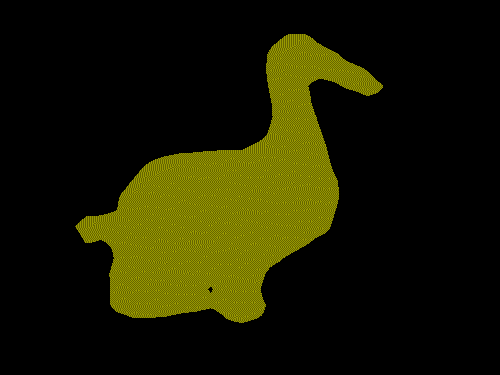} & 
			\includegraphics[width=.12\linewidth]{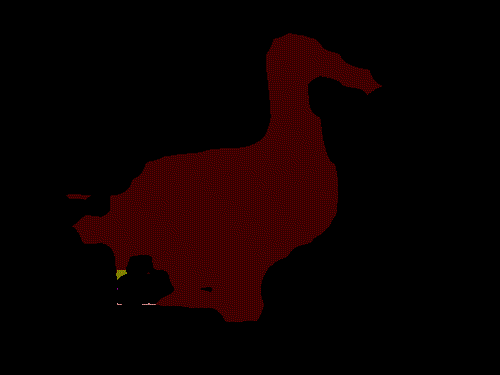} & 
			\includegraphics[width=.12\linewidth]{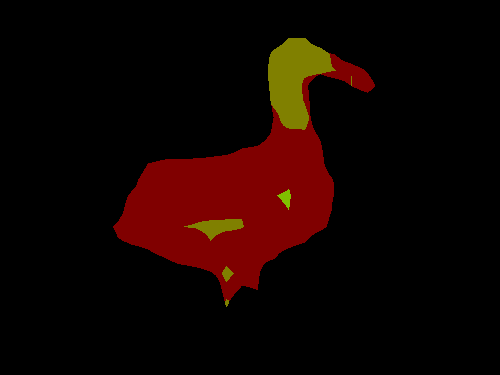} & 
			\includegraphics[width=.12\linewidth]{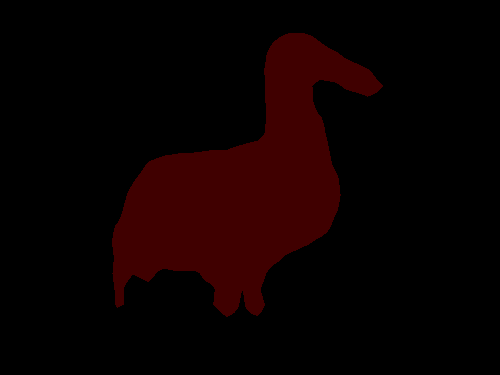}& 
			\includegraphics[width=.12\linewidth]{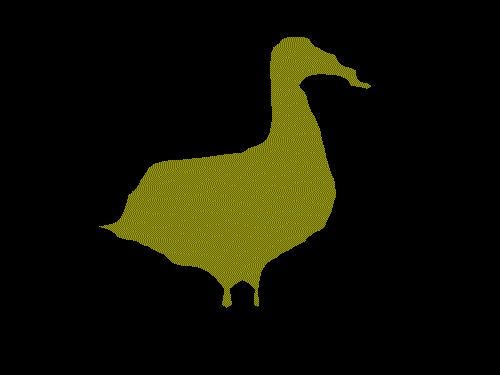}\\[-0.4ex]
			
			\includegraphics[width=.12\linewidth]{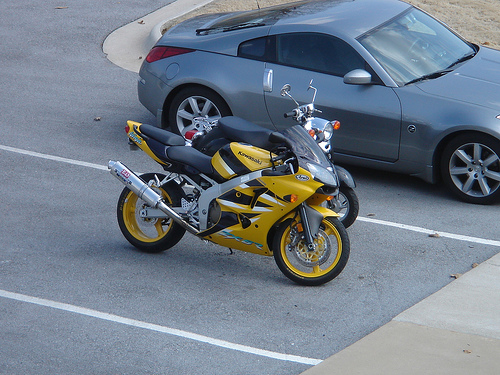} &
			\includegraphics[width=.12\linewidth]{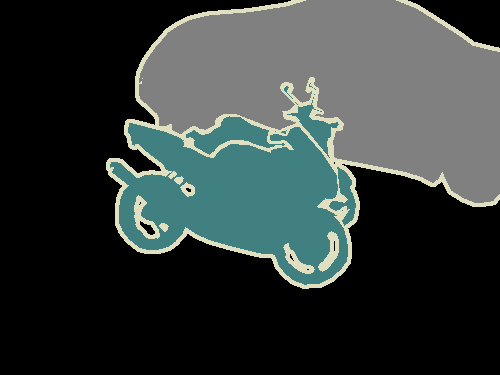}& 
			\includegraphics[width=.12\linewidth]{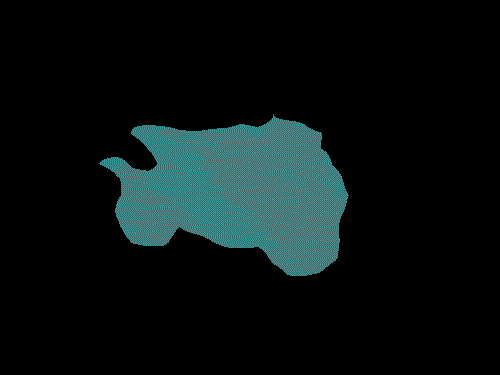}& 
			\includegraphics[width=.12\linewidth]{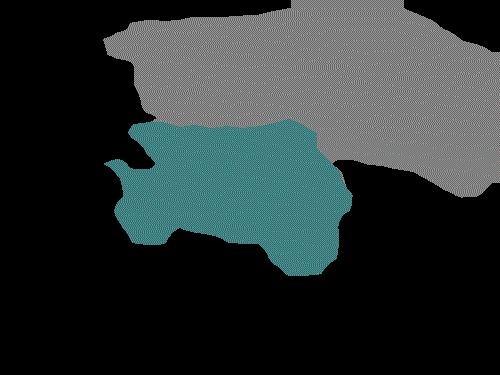} & 
			\includegraphics[width=.12\linewidth]{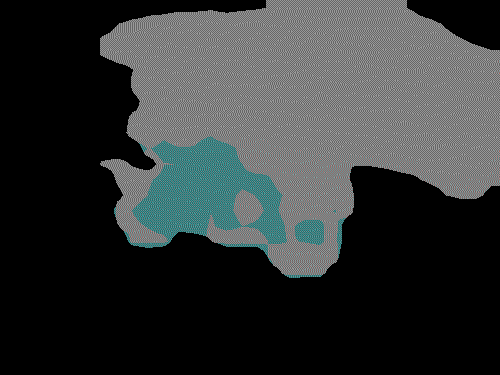} & 
			\includegraphics[width=.12\linewidth]{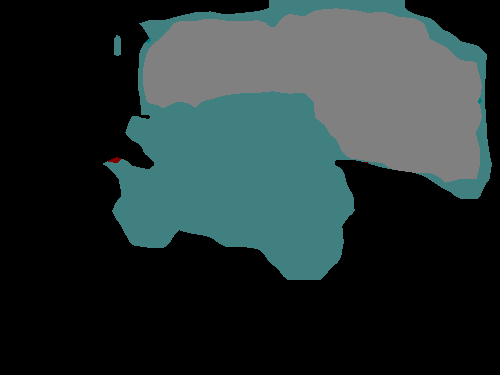} & 
			\includegraphics[width=.12\linewidth]{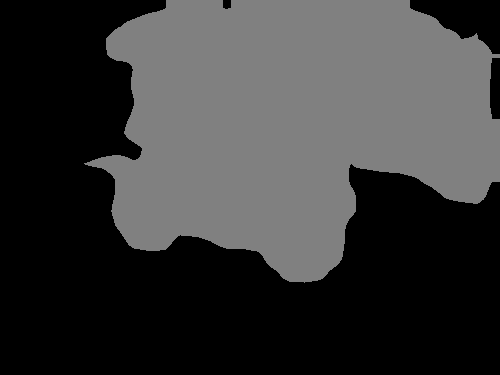}& 
			\includegraphics[width=.12\linewidth]{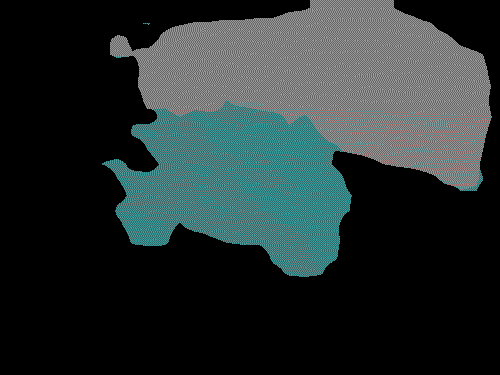}\\[-0.4ex]
			
			\multicolumn{8}{c}{\includegraphics[width=\linewidth]{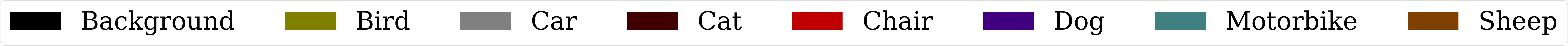}} \\
			
		\end{tabular}
	\end{adjustbox}
	\caption{Qualitative comparison on PASCAL VOC \cite{everingham2010pascal} \textit{VOC-Ex3}  after the final task.}
\label{fig:voc_ex3_viz}
\end{figure*}

The results on the three settings are shown in \cref{tab:pascal}, while the corresponding visualizations for \textit{VOC-Ex3} are presented in \cref{fig:voc_ex3_viz}. The performance on the unsplit classes, \textit{person} class in this dataset, remains consistent across the CL methods. We observe the biggest difference in performance between our approach and others for \textit{VOC-Ex1}. This is likely due to the more challenging multiple splitting of classes into intermediate subclasses. The lowest results for all methods are for \textit{VOC-Ex2}, which involves a longer sequence of tasks, and learning sets of mixed classes one from each parent class at each step. 

\subsubsection{Mapillary Vistas}

\begin{table*}[t]
	\centering
	\caption{Results for CLEO settings on Mapillary Vistas \cite{neuhold2017mapillary} after learning all tasks.} 
\begin{tabular}{c||cccc||cccc}
	\boldhline
	
	\multirow{2}{*}{\textbf{Method}} & \multicolumn{4}{c||}{\textbf{MV-Ex1}} & \multicolumn{4}{c}{\textbf{MV-Ex2}} \\ 
	& \textit{Unsplit} & \textit{Split} & \textit{Retained} & \textbf{All} & \textit{Unsplit} & \textit{Split} & \textit{Retained} & \textbf{All} \\ \boldhline
	
	Fine-Tuning & 00.00 & 18.59 & 00.00 & 08.69 & 00.00 & 01.47 & 00.00 & 00.69 \\
	\hline
	Joint Training & 28.37 & 22.68 & 25.79 & 25.51 & 28.37 & 22.68 & 25.79 & 25.51  \\
	\boldhline
	
	MiB \cite{MiB}& 27.02 & 16.33 & 17.74 & 21.27 & 09.83 & 04.31 & 00.07 & 06.45 \\ \hline
	
	PLOP \cite{PLOP}& 27.07 & 13.14 & 18.44 & 19.86 & 03.59 & 01.26 & 00.00 & 02.22  \\ \hline
	
	RCIL \cite{RCIL} & 28.54 & 01.57 & 12.78 & 14.65 & 06.39 & 00.55 & 00.17 & 03.16 \\ \hline
	
	MoOn (Ours) & 26.72 & 16.67 & 19.47 & \textbf{21.44} & 09.81  & 04.40 & 00.01 & \textbf{06.49} \\ \boldhline
	
\end{tabular}

\label{tab:mapillary}
\end{table*}

Using the two dataset versions of Mapillary Vistas \cite{neuhold2017mapillary}, we have defined two CLEO experiments. Both settings begin by learning the original 66 classes in task 0. The newly introduced classes are grouped into their ten parent classes from the previous version resulting in the original classes. The class mapping between the new classes and their parent classes is provided in the supplementary. The subsequent steps involve learning these new classes.

\begin{itemize}
\item \textbf{MV-Ex1:} After learning the original classes in task 0, all of the new classes are learned in task 1, resulting in two tasks. This exemplifies a real-world application of CLEO, where the model is specifically trained to learn the new classes without a complete retraining of all the classes.

\item \textbf{MV-Ex2:} This setting is more challenging with a longer sequence of tasks, which can exacerbate forgetting. In each step, one of the ten parent classes is split into its new classes, resulting in a total of eleven tasks.

\end{itemize}

Mapillary Vistas is challenging due to the large number of classes. 
As shown in \cref{tab:mapillary}, all methods exhibit suboptimal performance, including joint training which does not suffer from forgetting. 
The two settings are similar and have learned the same classes after the final task. They differ in the number of steps, highlighting the effect of a longer sequence of tasks. \textit{MV-Ex2} with a longer sequence significantly impacts the performance of all methods.

\section{Limitations and Future Work} \label{sec:limitations}
In its most general form, CLEO allows for \textit{merging} of two or more known classes as a valid form of evolution. 
However, we note that the here presented formalism of CLEO only considers the separation of known parent classes.
Yet, in some rare cases, it might be required to combine previous information to simplify previous tasks and free representational capacity of a model. 
This, however, as a version of \textit{intended forgetting} opens up a mostly unexplored field of research in continual learning and is left for future work.
Besides, semantic context can be achieved through post-processing by combining predictions of existing and new classes into a single class, preventing interference and avoiding further forgetting.

We further highlight that our baseline solution MoOn requires the information of class relations $S_t$ about the evolution of classes.
While we believe that this information is usually available to a sequentially trained model, there might be cases in which explicit knowledge about the ontology is unavailable. 
In such cases, $S_t$ could be inferred from the existing model's predictions on the current task data.
If the model predicts an existing class, the class being learned was previously seen and is split from this parent class. If it predicts the background, the new class is entirely new and unseen.
In our experimental evaluation, we found that MoOn and other approaches notably underperform with longer sequences in CLEO, highlighting the need for future attention to long-term learning settings.

\section{Conclusion}
We introduce a novel framework for addressing the challenge of adapting to evolving classes, termed Continual Learning of Evolving Ontologies (CLEO). 
The motivation for CLEO stems from a practical scenario wherein pre-existing classes undergo transformations over time, resulting in more specialized classes. 
CLEO extends the functionality of CIL to not just learn new information but also refine existing information. 
The applicability of CLEO in the real world is exemplified through the Mapillary Vistas dataset, where an existing set of classes evolves into more detailed classes. 
Furthermore, we highlight the shortcomings of existing CIL methods and propose MoOn as an initial solution for CLEO. The effectiveness of MoOn for CLEO is demonstrated on seven experimental settings across three datasets, surpassing previous CIL approaches. 

\section*{Acknowledgments}
This work was partially funded by the Federal Ministry of Education and Research Germany under the project DECODE (01IW21001).

\newpage

\markright{CLEO: Supplementary Material}

\begin{appendix}
{
    \centering
    \Large
    \bf
    Supplementary Material
    \vskip .25em
}

\section{Overview}
This document provides the supplementary material to our main paper on \textit{CLEO: Continual Learning of Evolving Ontologies}.
We first present the detailed class hierarchies that we use to design our experiments for Cityscapes \cite{cordts2016cityscapes}, PASCAL VOC \cite{everingham2010pascal}, and Mapillary Vistas \cite{neuhold2017mapillary}.
Afterwards, we extend our motivation for the design of these experiments, \ie explain how they differ and which possible types of evolution are covered in each experiment.
Additionally, we present the task-wise evaluation of our approach MoOn and the state-of-the-art for Cityscapes and provide a visualization for the task-wise evolution.
The task-wise evaluation is extended by a detailed breakdown of per-class results after each task.
Finally, we list the exact sets of classes for each task of each experiment as well as the final groups of classes that are considered in our evaluation of CLEO.

\section{Class Hierarchies}

\subsection{Cityscapes}

\begin{figure}[t] 
	\centering
	\includegraphics[width=0.45\linewidth]{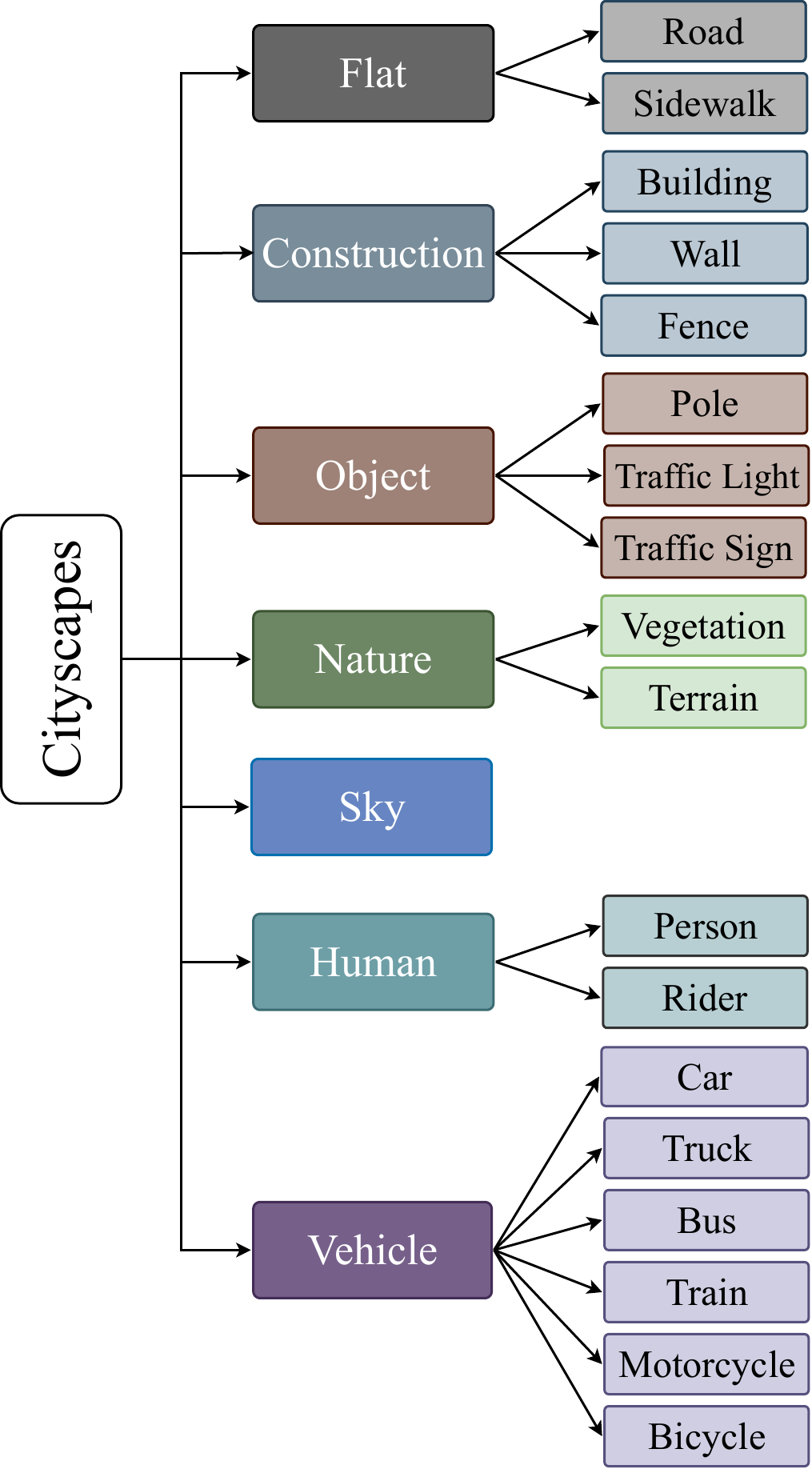} \caption{The official label hierarchy of Cityscapes \cite{cordts2016cityscapes}.}
	\label{fig:cs_hierarchy}
\end{figure}

The Cityscapes dataset \cite{cordts2016cityscapes} provides annotations for 30 classes categorized into 8 groups namely, \textit{flat, construction, object, nature, sky, human,} and \textit{void} classes.
Within these 30 classes, 19 classes are used for training and evaluation, while the remainder are treated as void.
These 19 classes and their grouping into the 7 parent classes are visualized in \cref{fig:cs_hierarchy}.
Notably, \textit{sky} is the only class which in turn does not contain any further classes.
The dataset exhibits varying levels of representation among the individual classes, with some being underrepresented, which significantly impacts the learning process.
We use this class hierarchy to derive two settings for the Cityscapes dataset and list the classes included in each task in \cref{tab:app:csex1,tab:app:csex2}.

\subsection{PASCAL VOC 2012}

\begin{figure}[t] 
	\centering
	\includegraphics[width=0.6\linewidth]{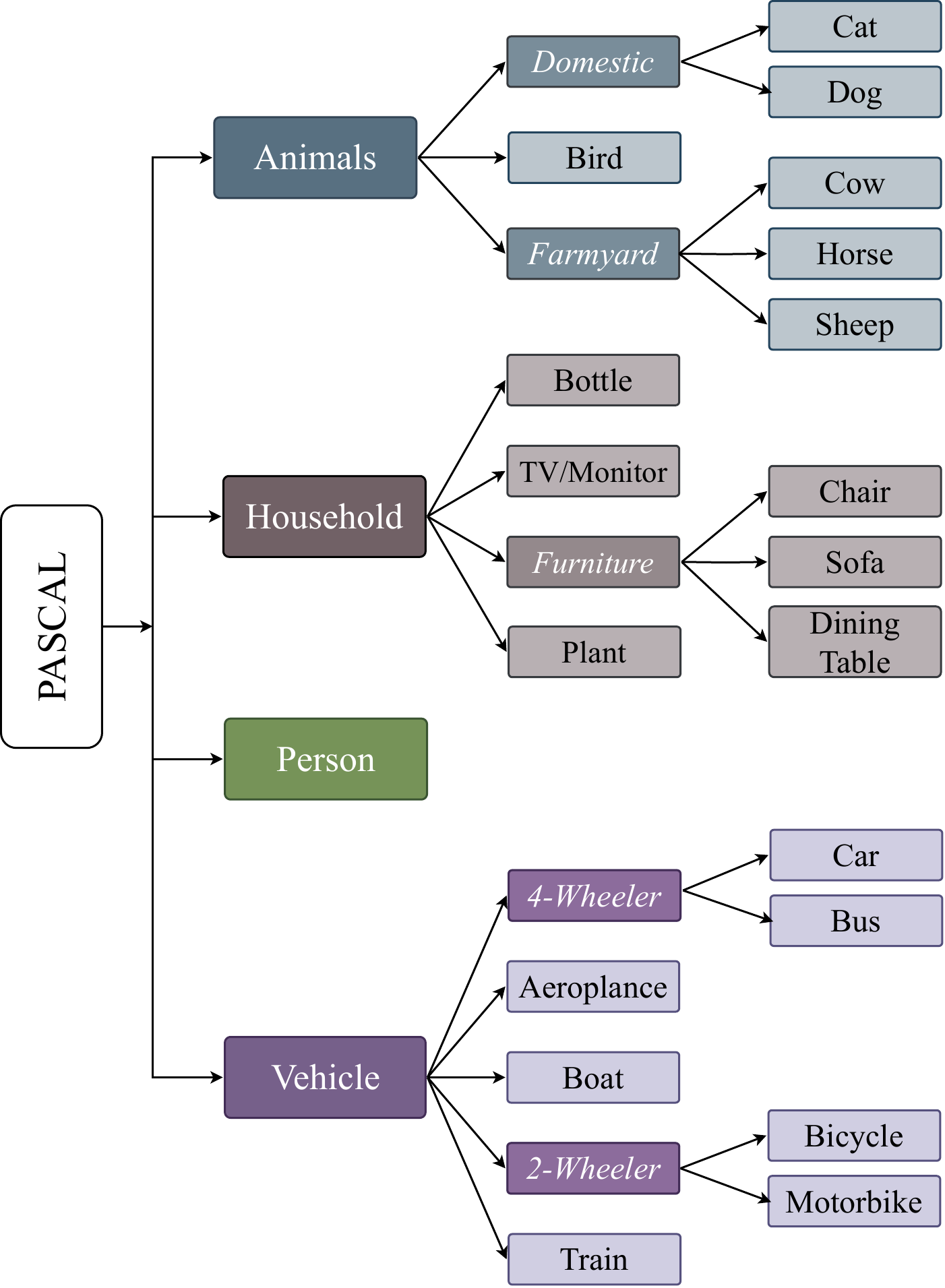}
    \caption{Label hierarchy for PASCAL VOC \cite{everingham2010pascal} adapted from \cite{VOC_Hierarchy}.}
	\label{fig:PASCAL}
\end{figure}

The PASCAL VOC 2012 dataset \cite{everingham2010pascal} is widely used for benchmarking tasks such as classification, object detection, and semantic segmentation.
It provides annotations for 20 object categories including common objects.
A label hierarchy for PASCAL VOC is provided by \cite{VOC_Hierarchy}, which groups these 20 classes into 4 parent classes with \textit{person} being the only class without any sub-classes.
Interestingly, the suggested hierarchy for PASCAL VOC consists of intermediate sub-groupings of these 20 classes into groups such as \textit{domestic animals, farmyard animals, furniture, 4-wheeler,} and \textit{2-wheeler}, resulting in a more challenging multi-level splitting and retention.
The class hierarchy representing the grouping of the 20 original classes at different levels, including their sub-groups is presented in \cref{fig:PASCAL}.
The task-wise evolution for the experiments on PASCAL VOC is given in \cref{tab:app:vocex1,tab:app:vocex2,tab:app:vocex3}.

\subsection{Mapillary Vistas}

\begin{figure}[ph!]
	\centering
	
	\hspace*{\fill}%
	\begin{subfigure}[c]{0.49\linewidth}
		\includegraphics[width=\linewidth]{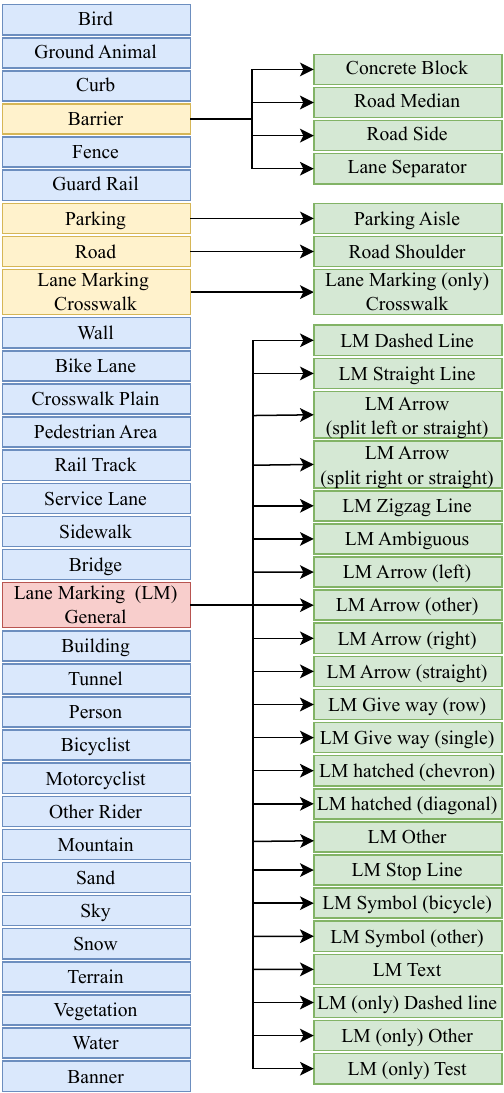}
	\end{subfigure}\hspace*{\fill}%
	\begin{subfigure}[c]{0.49\linewidth}
		\includegraphics[width=\linewidth]{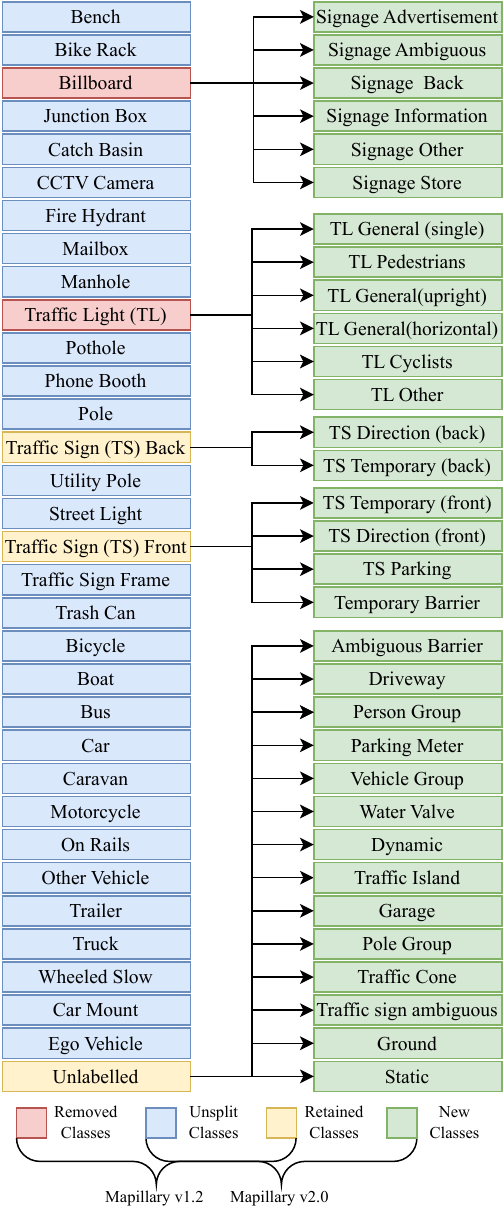}
	\end{subfigure}%
	\hspace*{\fill}
	
	\caption{Class mapping for between both versions of Mapillary Vistas \cite{neuhold2017mapillary}.}

\label{fig:MapillaryMapping}
\end{figure}

The initial version v1.2 of the Mapillary Vistas dataset \cite{neuhold2017mapillary} released in 2017 provided annotations for 66 object classes, which was later refined into 124 classes in the current version v2.0.
The Mapillary Vistas dataset does not provide an official mapping between the two versions.
However, since both versions of the dataset annotate the same set of images, we can create a class mapping by comparing the ground truth of both versions to determine how the classes have changed. The class mapping is visualized in \cref{fig:MapillaryMapping}.
Through this process, we have identified ten parent classes in the initial version, from which the newly defined classes in v2.0 have originated.
These classes include: \textit{barrier, lane marking - crosswalk, parking, road, traffic sign (front), traffic sign (back), unlabelled, billboard, lane marking - general, and traffic light}.
Version 2.0 introduces 14 entirely new classes that were previously grouped under the \textit{unlabelled} class, while the remaining have been split from existing classes to represent more specific semantics.
Notably, the classes, \textit{billboard, lane marking - general, and traffic light} have been completely split into new classes and are no longer part of the current version, while the other parent classes have been split up only partially.

\section{Differences between the Experiments}
In this section, we describe the variations within our seven different experiments.
The most prominent dimension along we vary is the dataset itself.
This variation implies application domains, numbers of classes, different sensors, \etc.
Our experiments on Cityscapes \cite{cordts2016cityscapes} cover two potential cases.
Firstly, in \textit{CS-Ex1}, the semantic evolution affects multiple superclasses in parallel, \ie during the same task.
Secondly, within the same task, multiple subclasses are split from the same superclass in \textit{CS-Ex2}.
As such, these two experiments cover varying numbers of involved super- and subclasses.

The applied label hierarchy of PASCAL VOC \cite{everingham2010pascal} adds an additional level of semantic resolution (\cf \cref{fig:PASCAL}), which allows us to design experiments in which splitting and retention can happen more than once, \ie a single class can undergo two evolutionary steps.

Lastly, with the vast number of classes in Mapillary Vistas \cite{neuhold2017mapillary}, we can cover two more use-cases and can define more challenging and complex scenarios.
\Ie, in \textit{MV-Ex1} we cover the realistic semantic evolution that was introduced by the authors of the dataset, that affects many super- and subclasses in a single task.
Due to the large number of classes, we can also design an experiment that covers a much longer sequence of tasks in \textit{MV-Ex2}, \ie 10 evolutionary steps.

\section{Task-wise Results}
\begin{figure}[ph!]
	\centering
	\setlength{\tabcolsep}{2pt}
	\large
	\begin{adjustbox}{width=\textwidth, totalheight=0.91\textheight}
		
		\begin{tabular}{cccc}
			
			\multirow{-5}{*}{\rotatebox[origin=c]{90}{\small Image}} \quad & 
			\includegraphics[width=.32\linewidth]{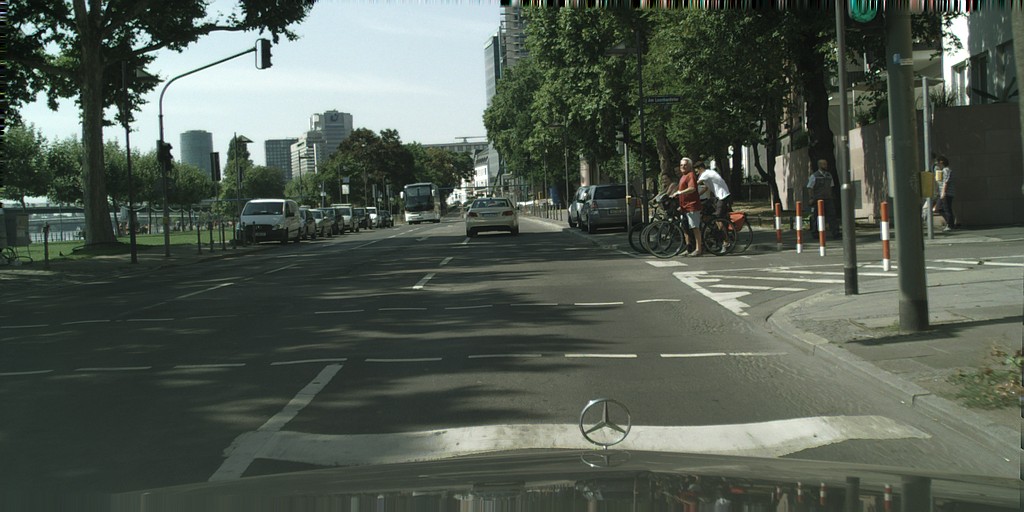} & \includegraphics[width=.32\linewidth]{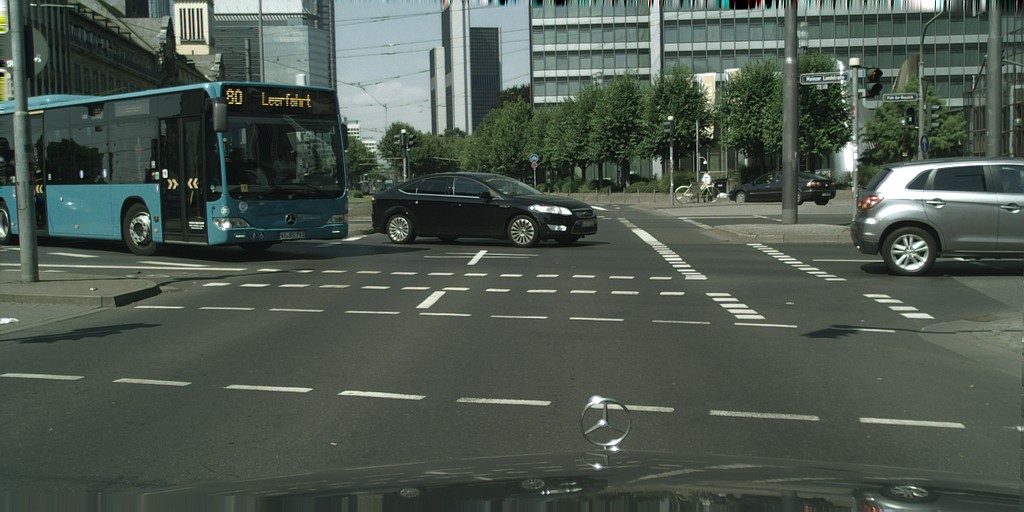}& \includegraphics[width=.32\linewidth]{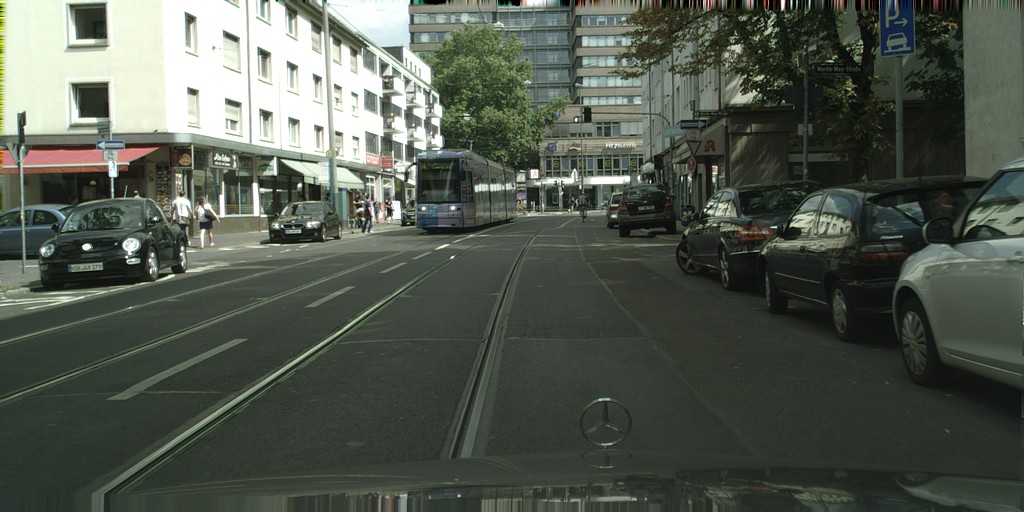}\\
			
			\multirow{-5}{*}{\rotatebox[origin=c]{90}{\small Task 0}} \quad & 
			\includegraphics[width=.32\linewidth]{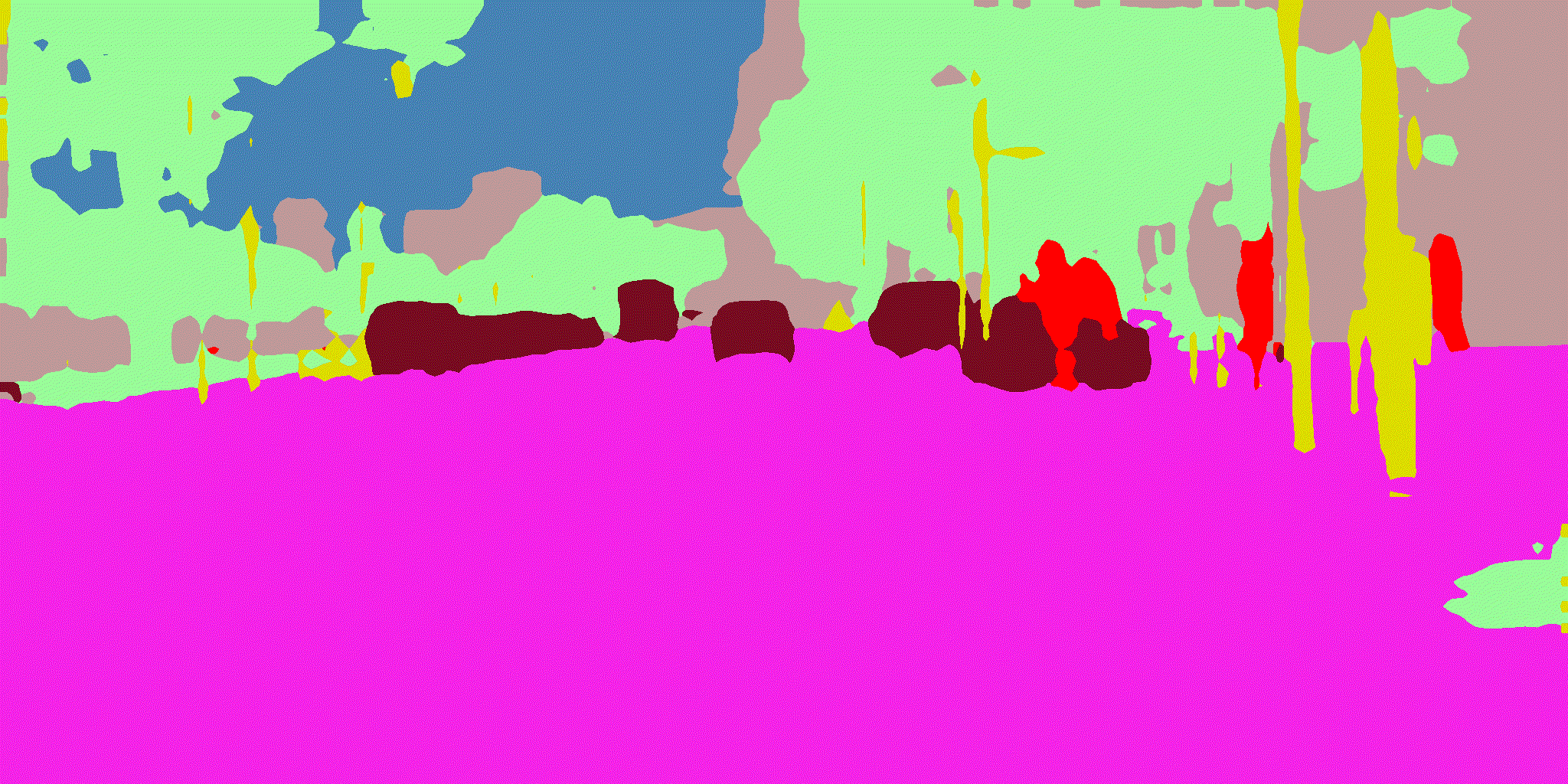} & \includegraphics[width=.32\linewidth]{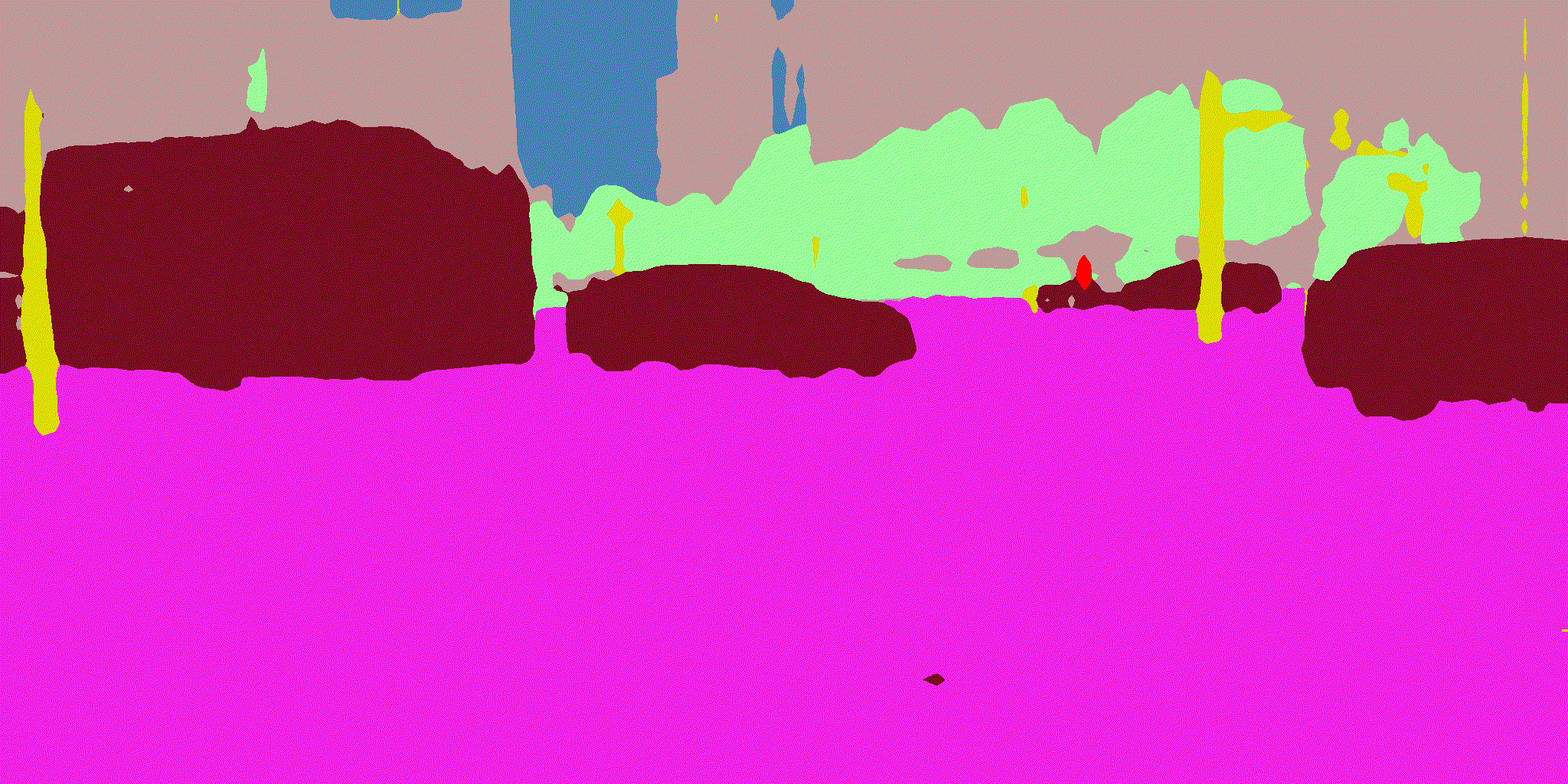}& \includegraphics[width=.32\linewidth]{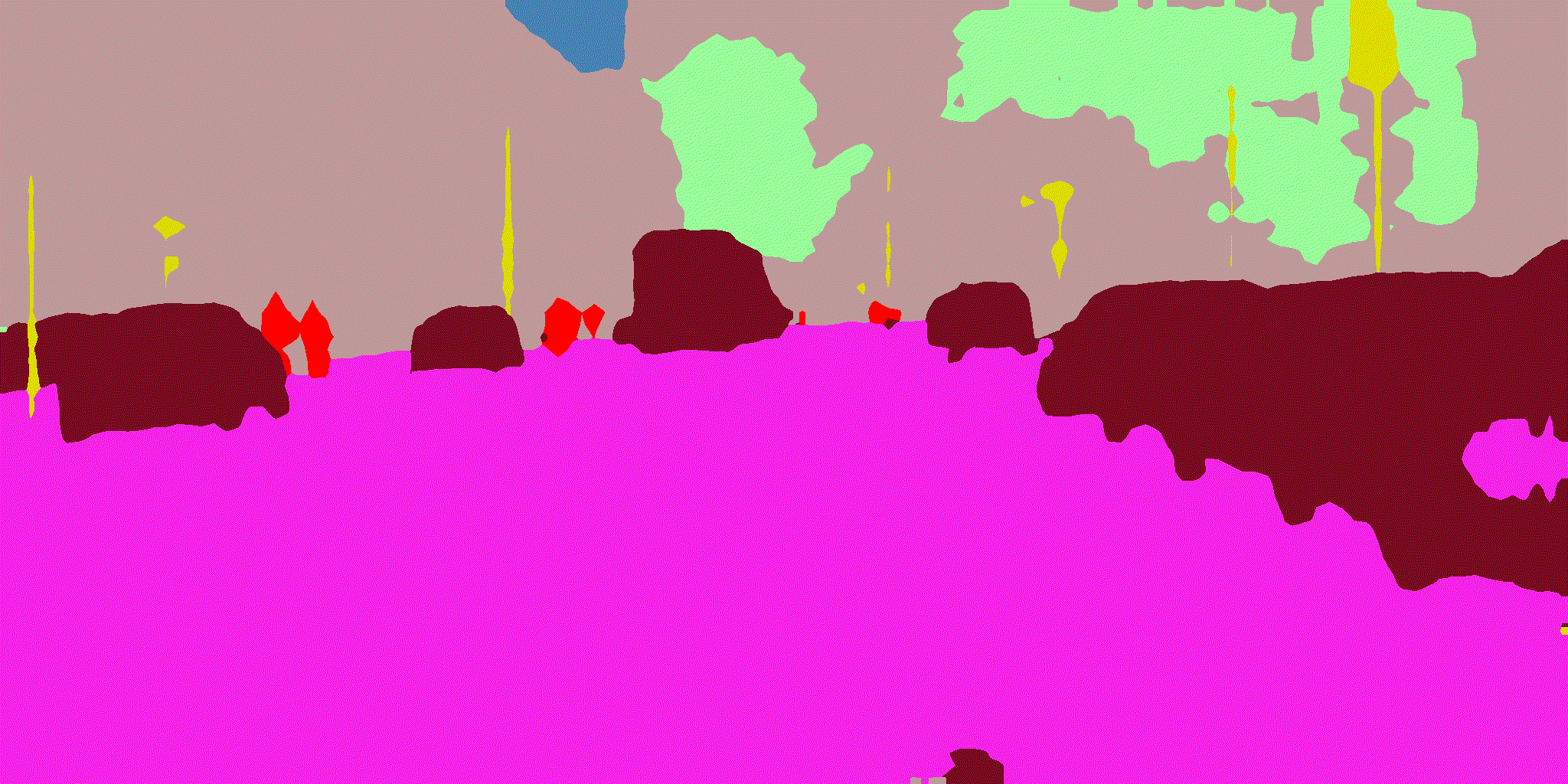}\\    
			
			\multirow{-5}{*}{} \quad & 
			\multicolumn{3}{c}{\includegraphics[width=0.9\linewidth]{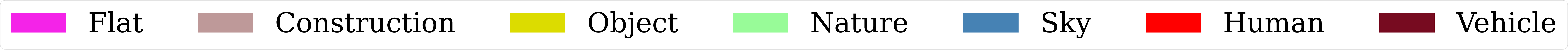}} \\
						
			\multirow{-5}{*}{\rotatebox[origin=c]{90}{\small Task 1}} \quad & 
			\includegraphics[width=.32\linewidth]{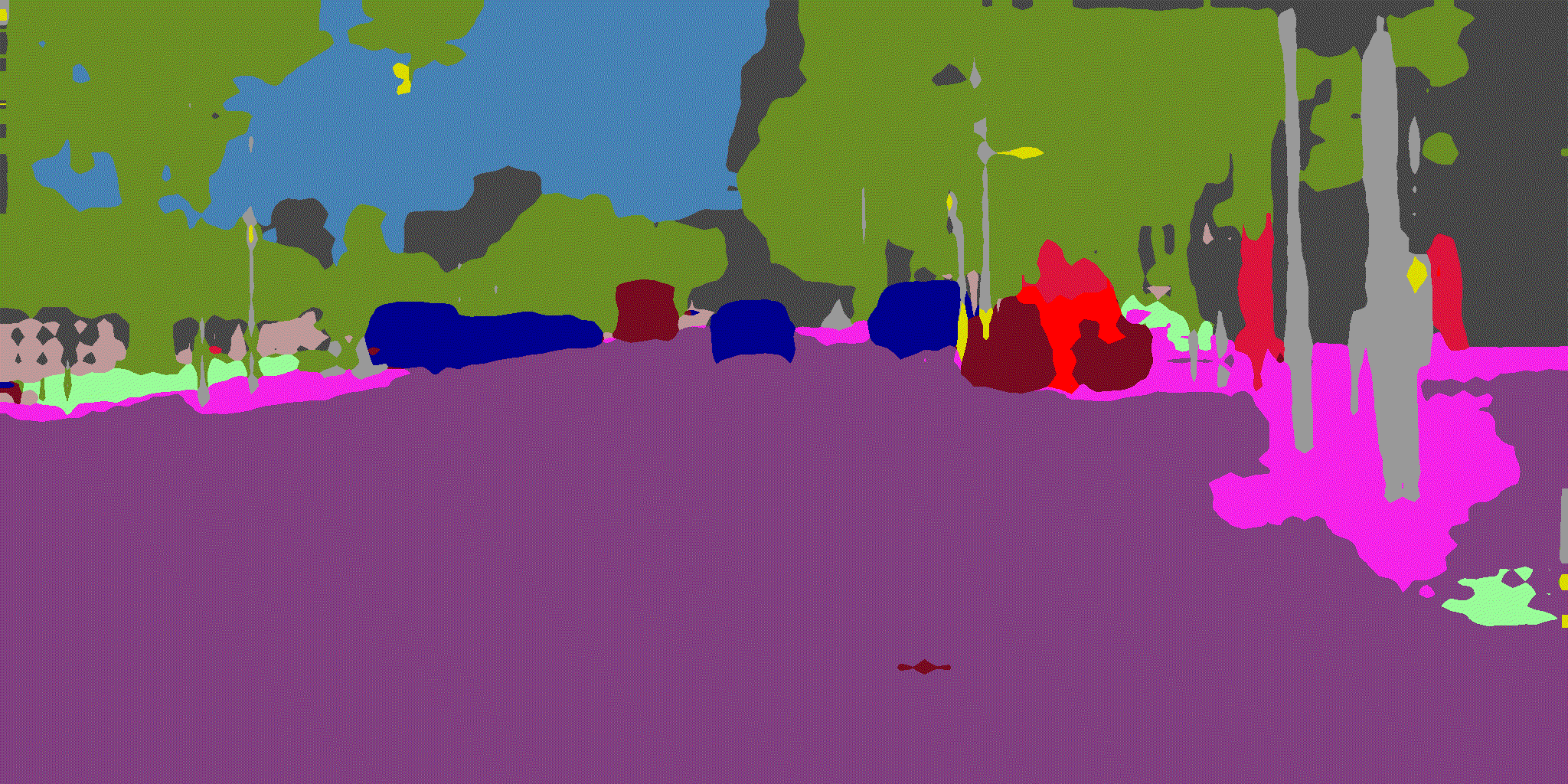} & \includegraphics[width=.32\linewidth]{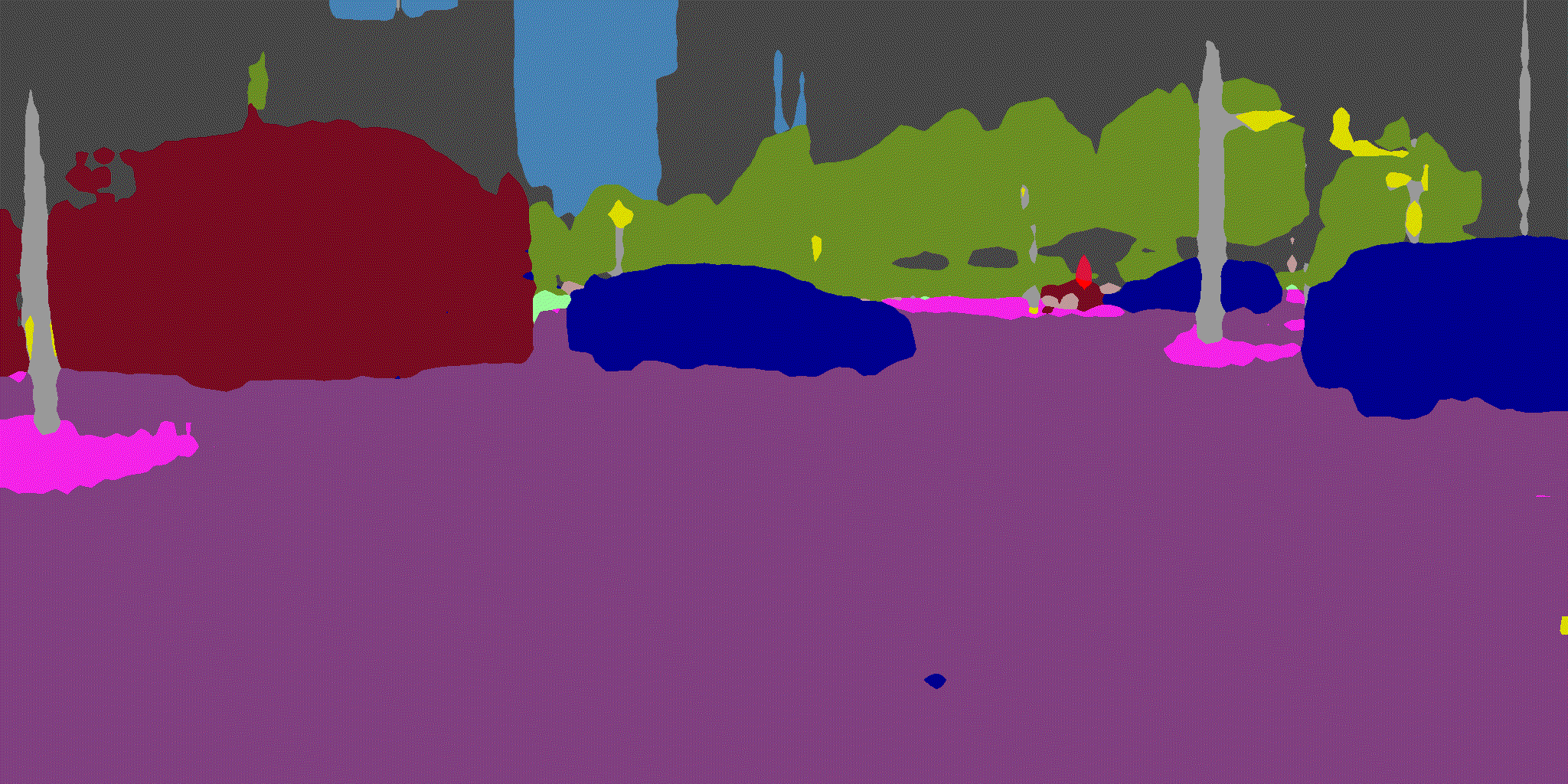}& \includegraphics[width=.32\linewidth]{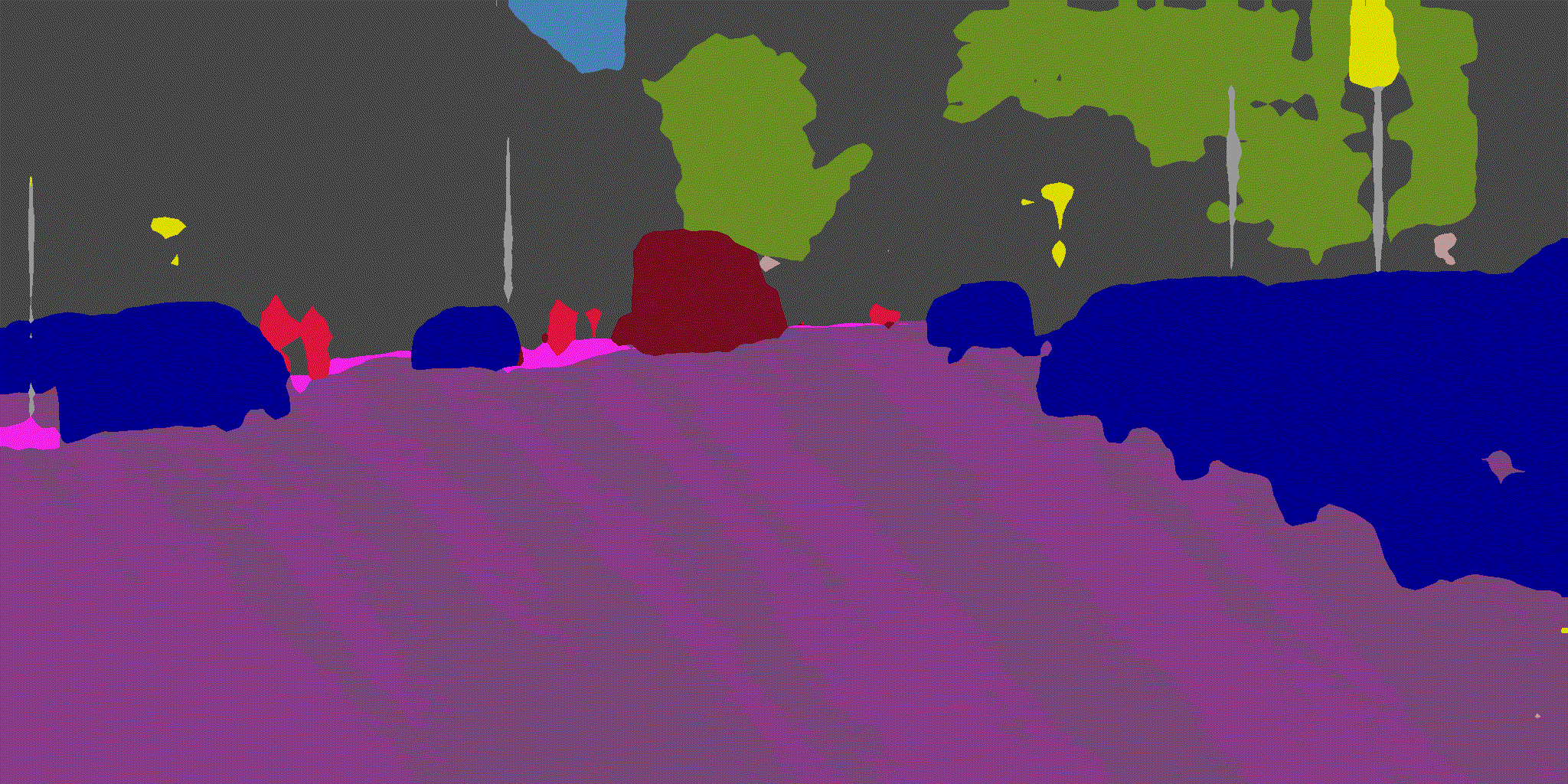}\\    
			
			\multirow{-5}{*}{} \quad & 
			\multicolumn{3}{c}{\includegraphics[width=0.9\linewidth]{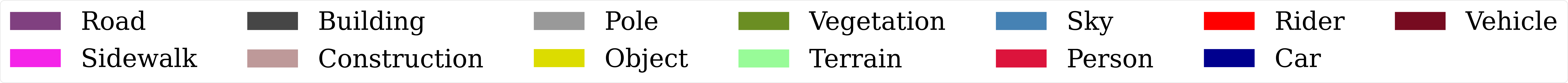}} \\
			
			\multirow{-5}{*}{\rotatebox[origin=c]{90}{\small Task 2}} \quad & 
			\includegraphics[width=.32\linewidth]{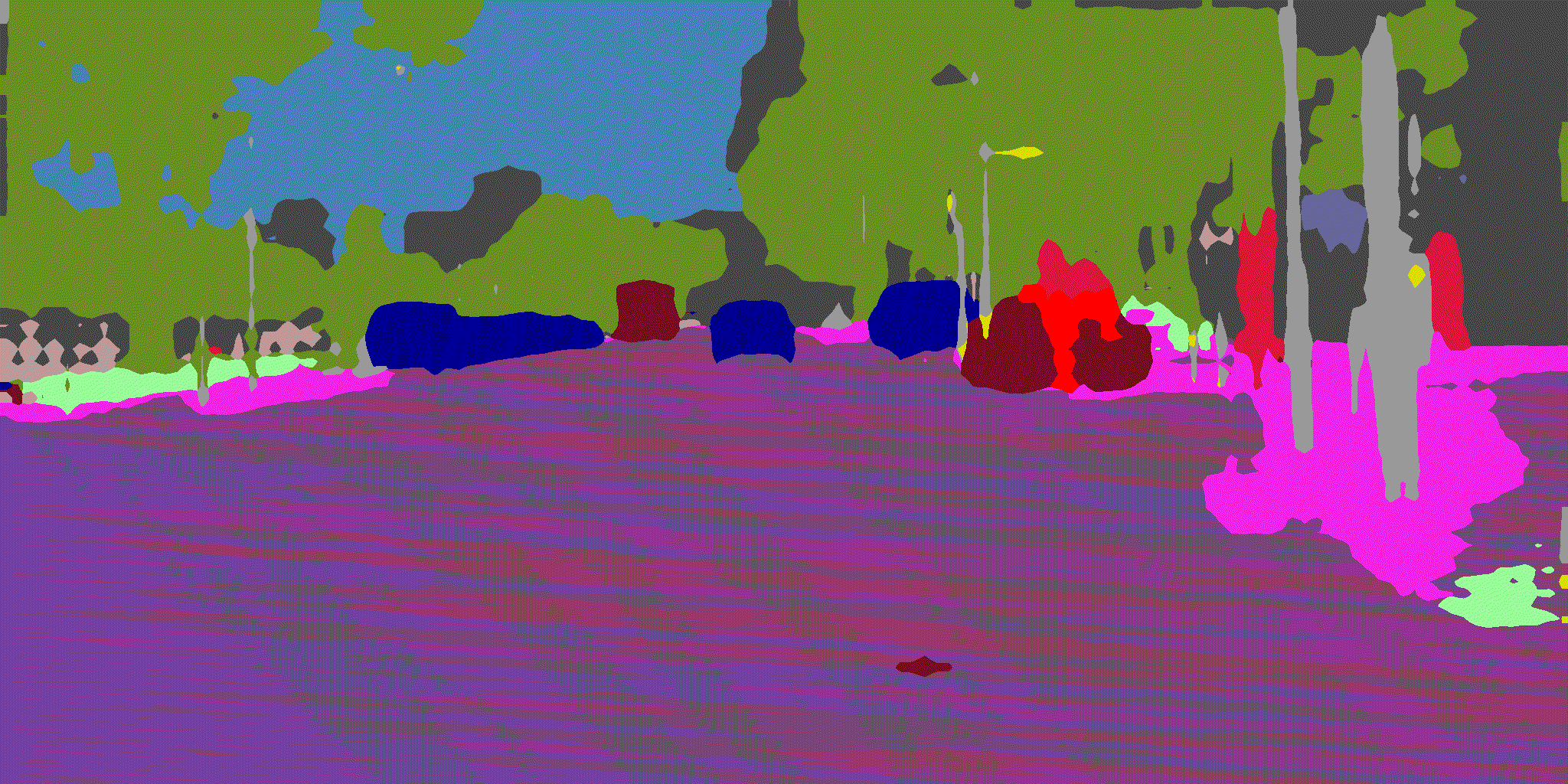} & \includegraphics[width=.32\linewidth]{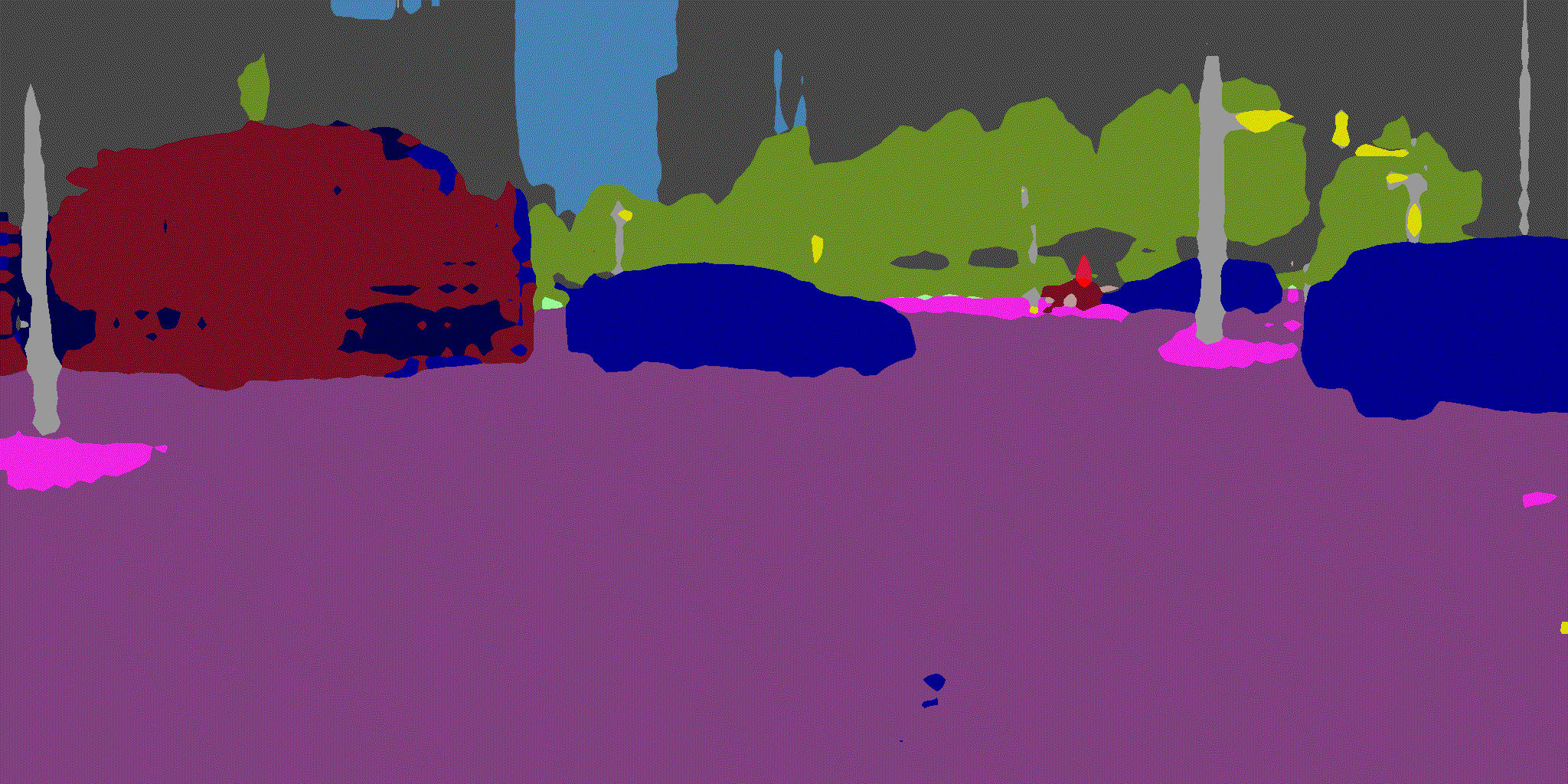}& \includegraphics[width=.32\linewidth]{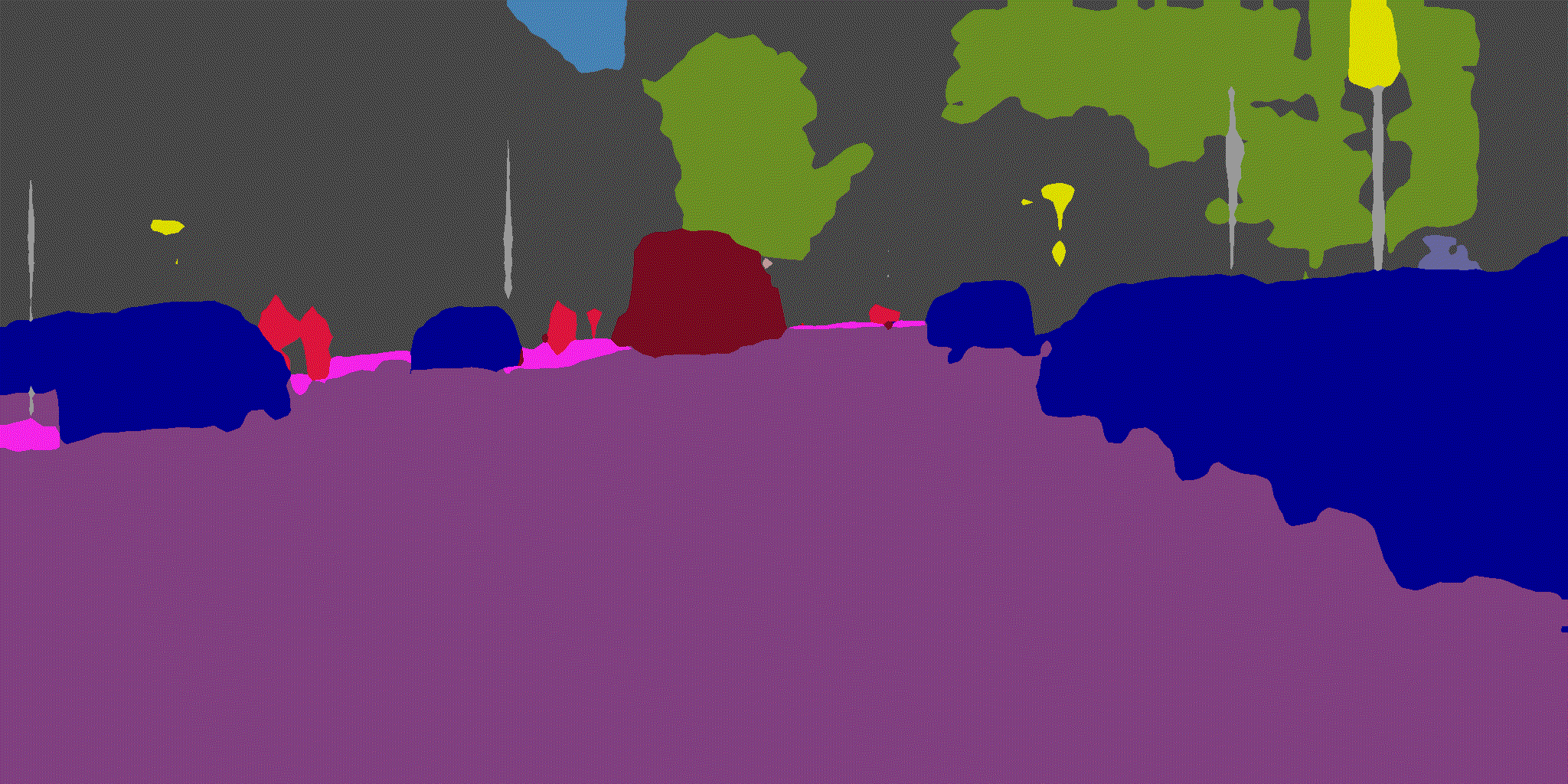}\\
			
			\multirow{-5}{*}{} \quad & 
			\multicolumn{3}{c}{\includegraphics[width=0.9\linewidth]{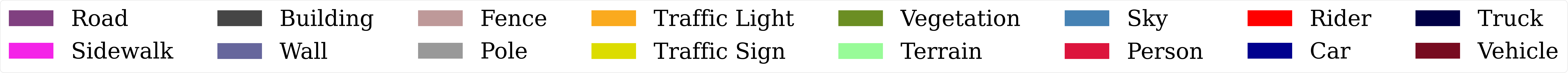}} \\    
			
			\multirow{-5}{*}{\rotatebox[origin=c]{90}{\small Task 3}} \quad & 
			\includegraphics[width=.32\linewidth]{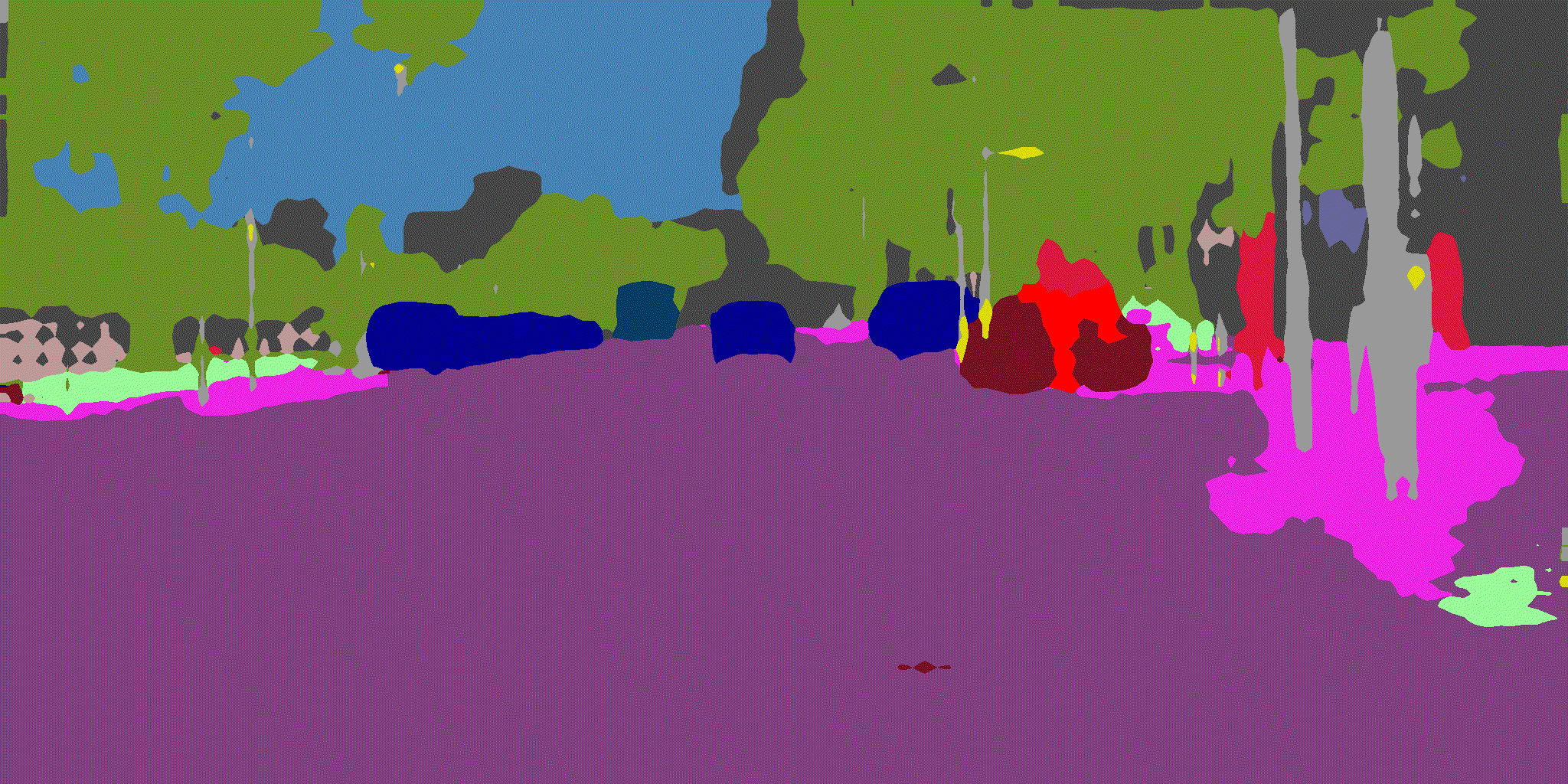} & \includegraphics[width=.32\linewidth]{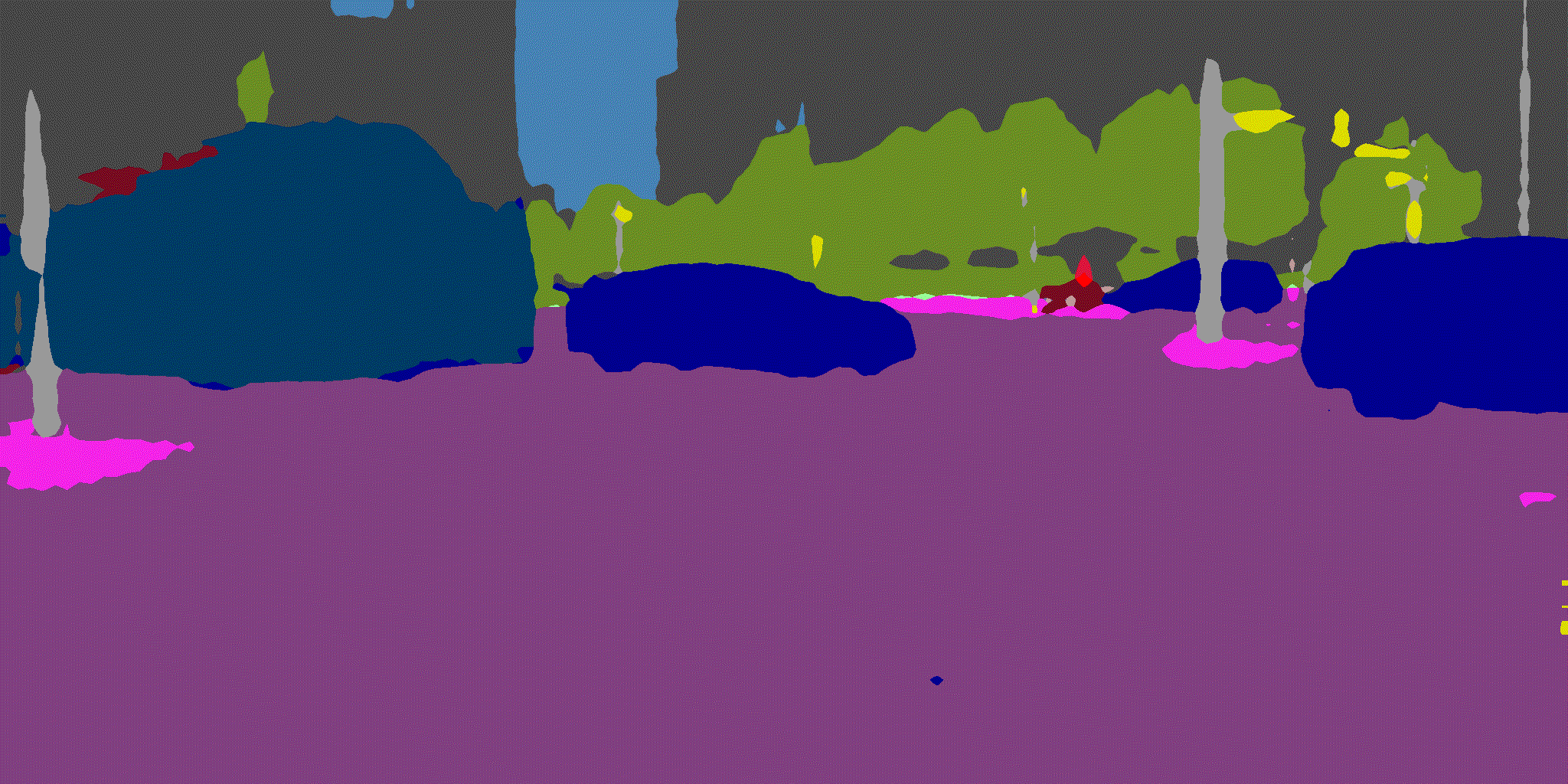}& \includegraphics[width=.32\linewidth]{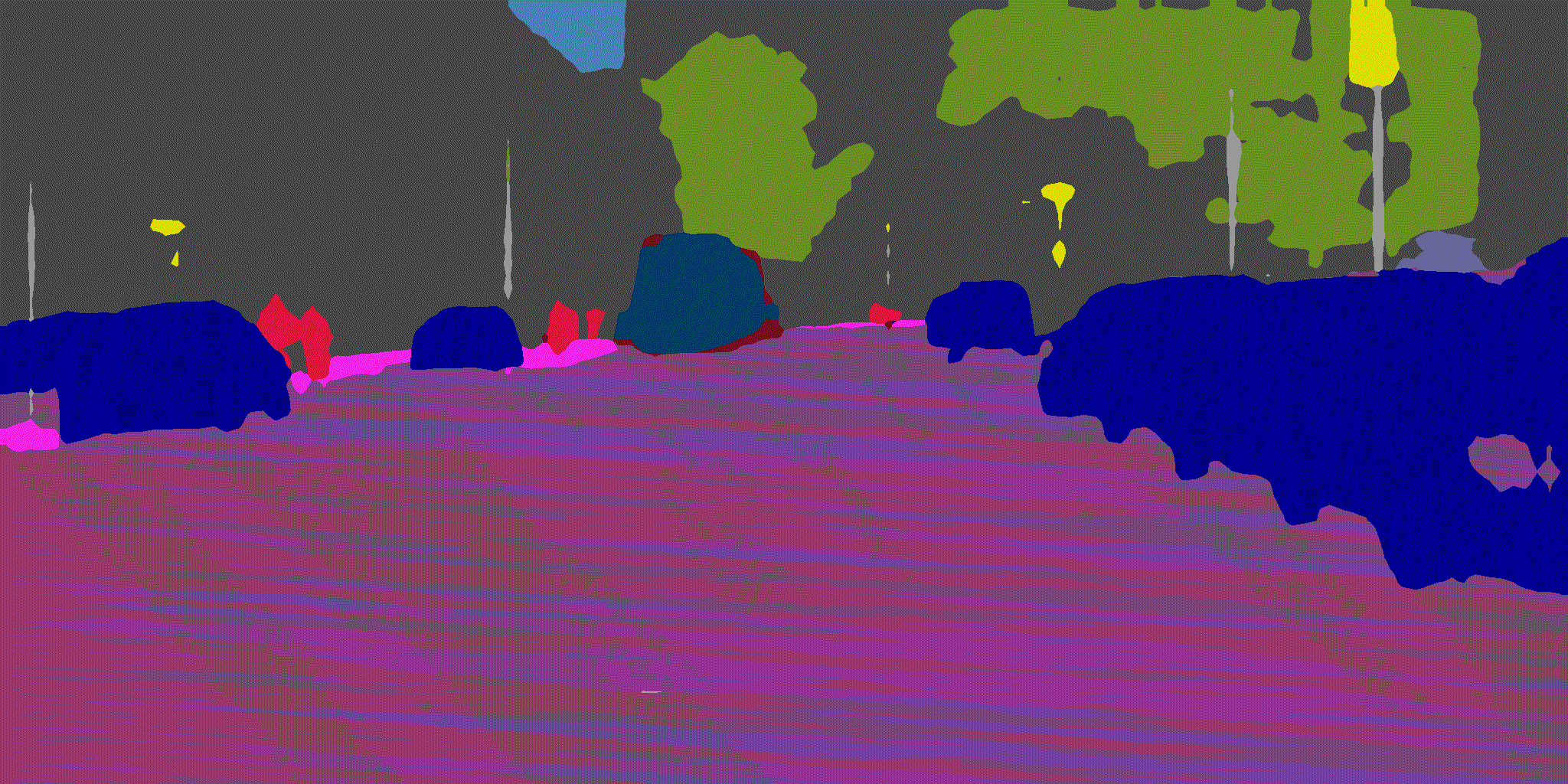}\\    
			
			\multirow{-5}{*}{} \quad & 
			\multicolumn{3}{c}{\includegraphics[width=0.9\linewidth]{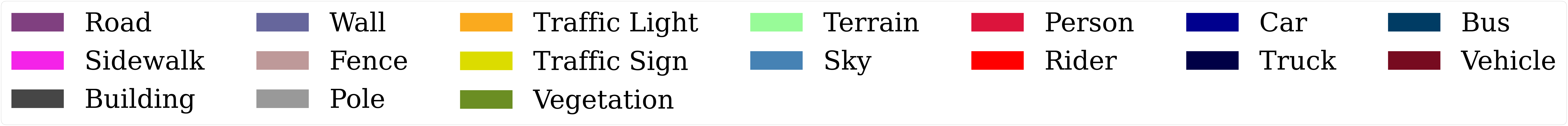}} \\
			
			\multirow{-5}{*}{\rotatebox[origin=c]{90}{\small Task 4}} \quad & 
			\includegraphics[width=.32\linewidth]{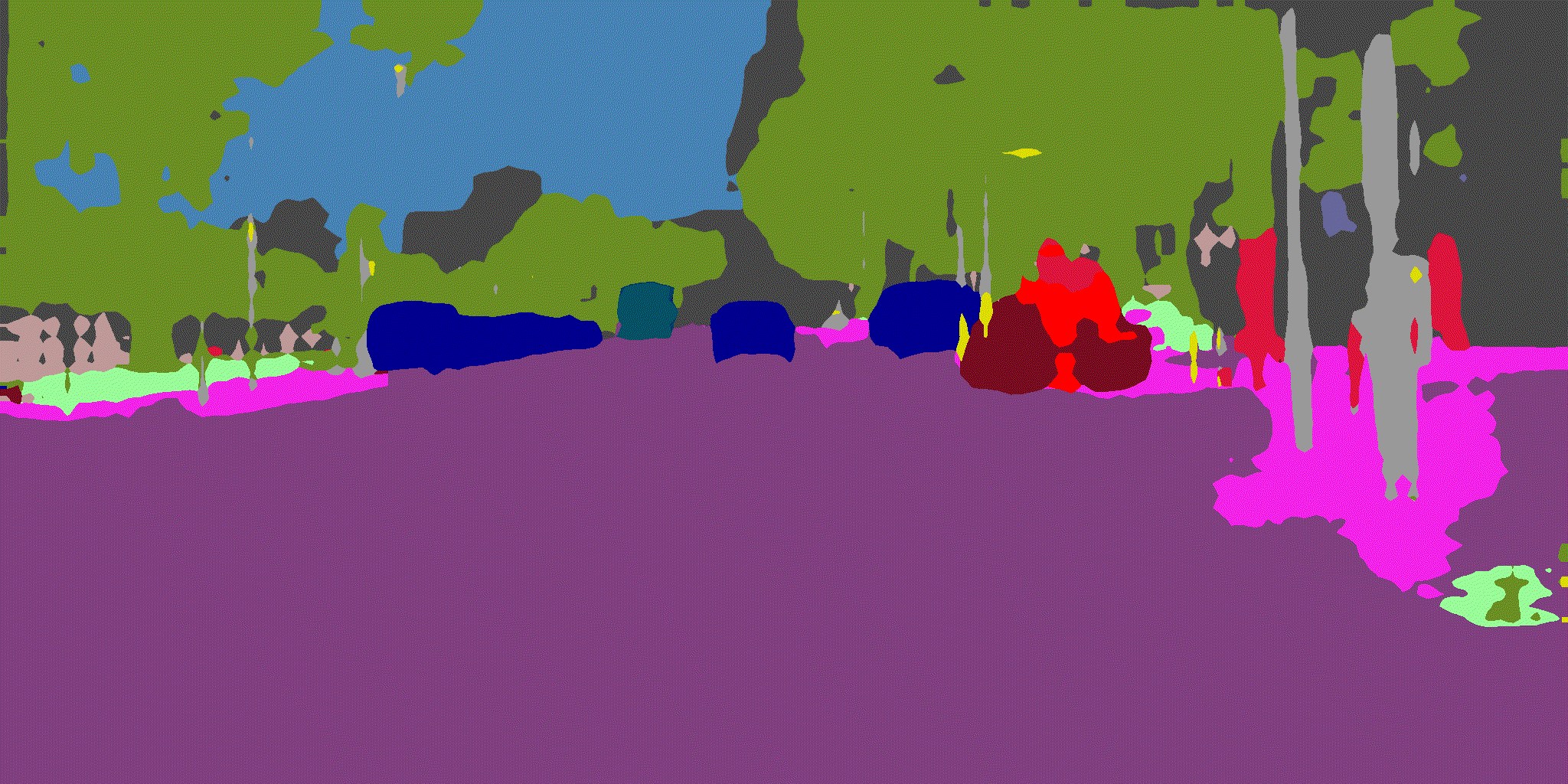} & \includegraphics[width=.32\linewidth]{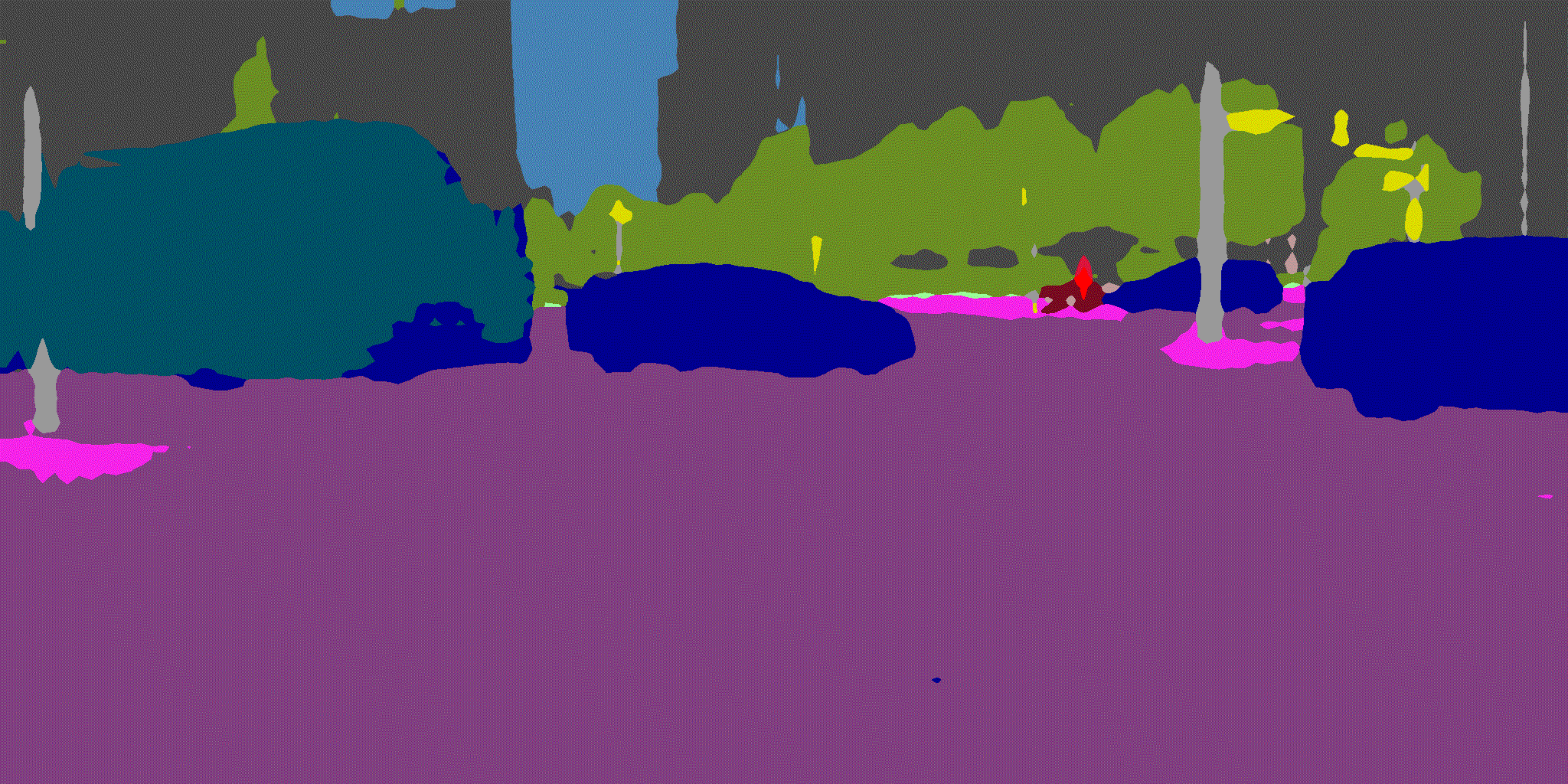}& \includegraphics[width=.32\linewidth]{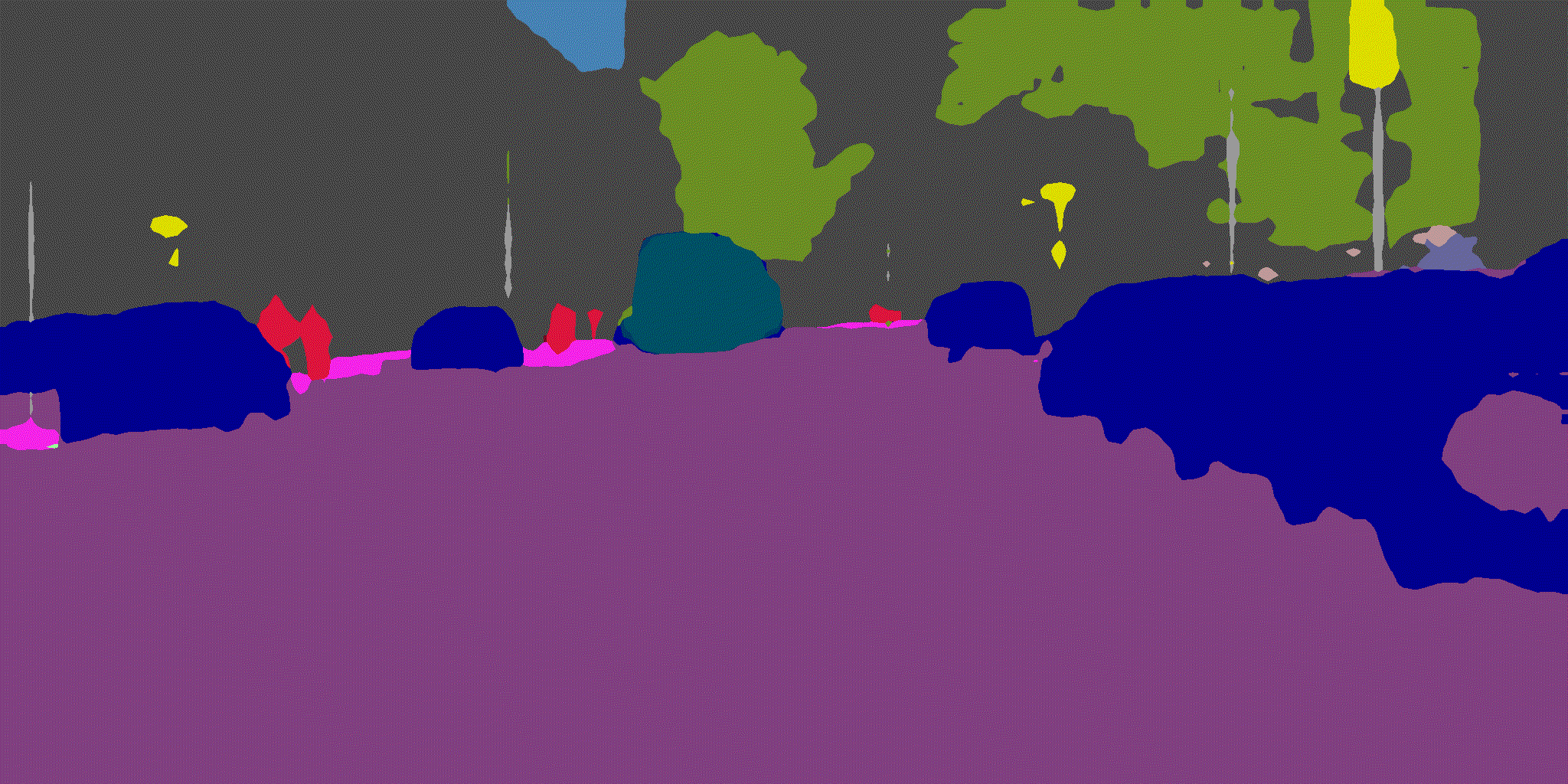}\\    
			
			\multirow{-5}{*}{} \quad & 
			\multicolumn{3}{c}{\includegraphics[width=0.9\linewidth]{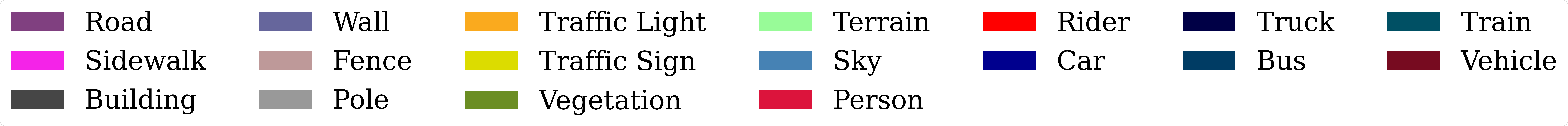}} \\
			
			\multirow{-5}{*}{\rotatebox[origin=c]{90}{\small Task 5}} \quad & 
			\includegraphics[width=.32\linewidth]{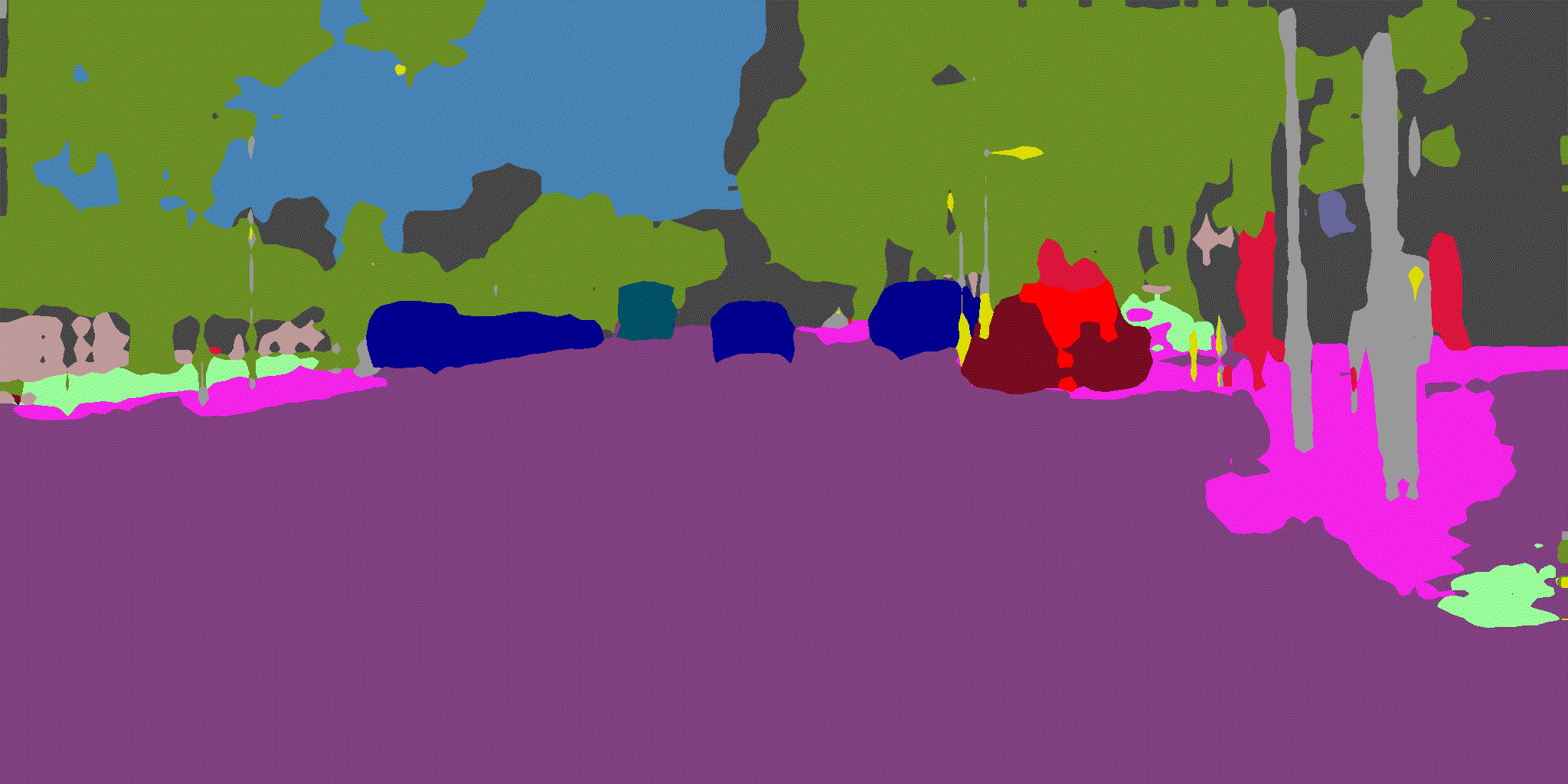} & \includegraphics[width=.32\linewidth]{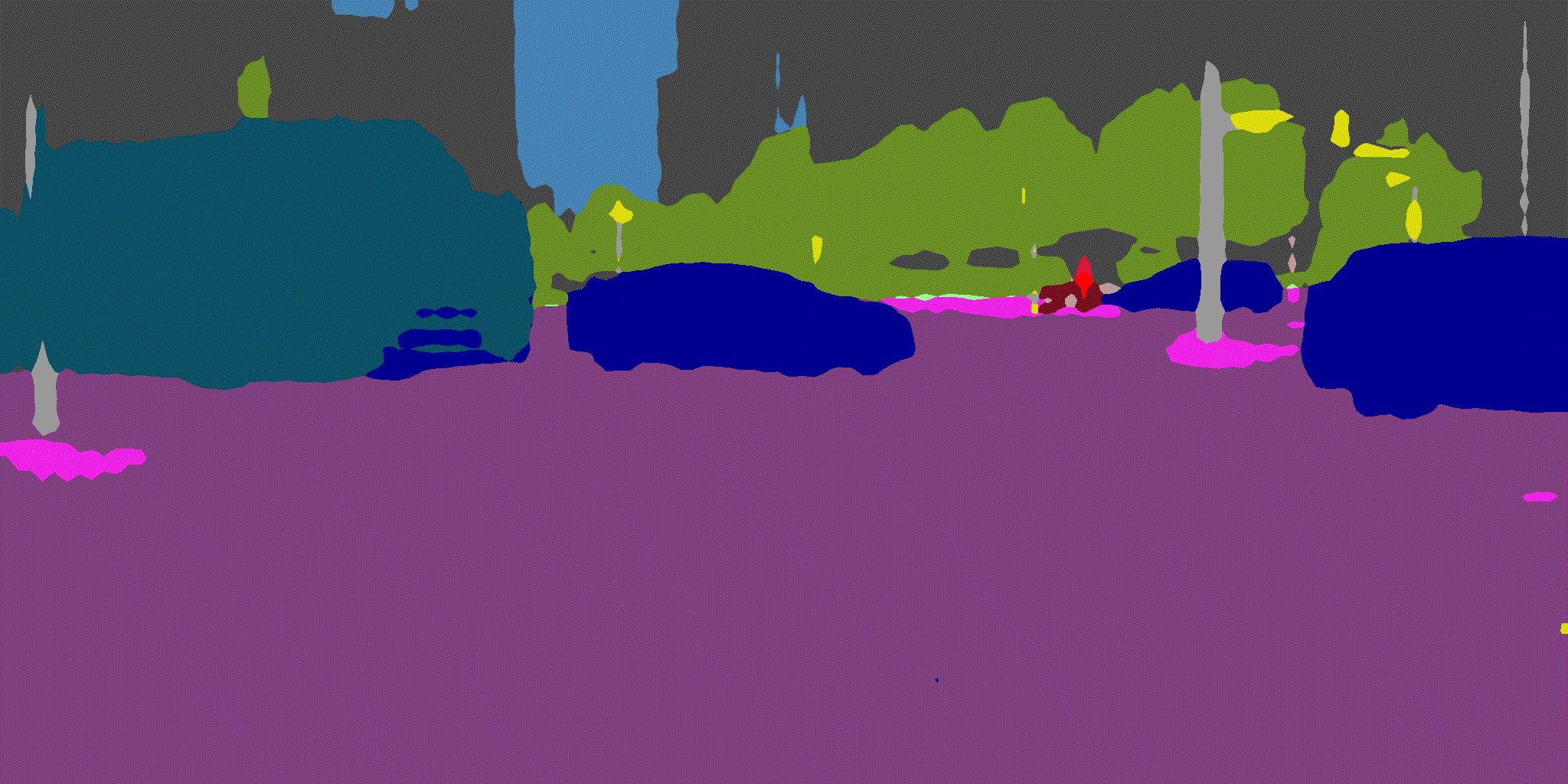}& \includegraphics[width=.32\linewidth]{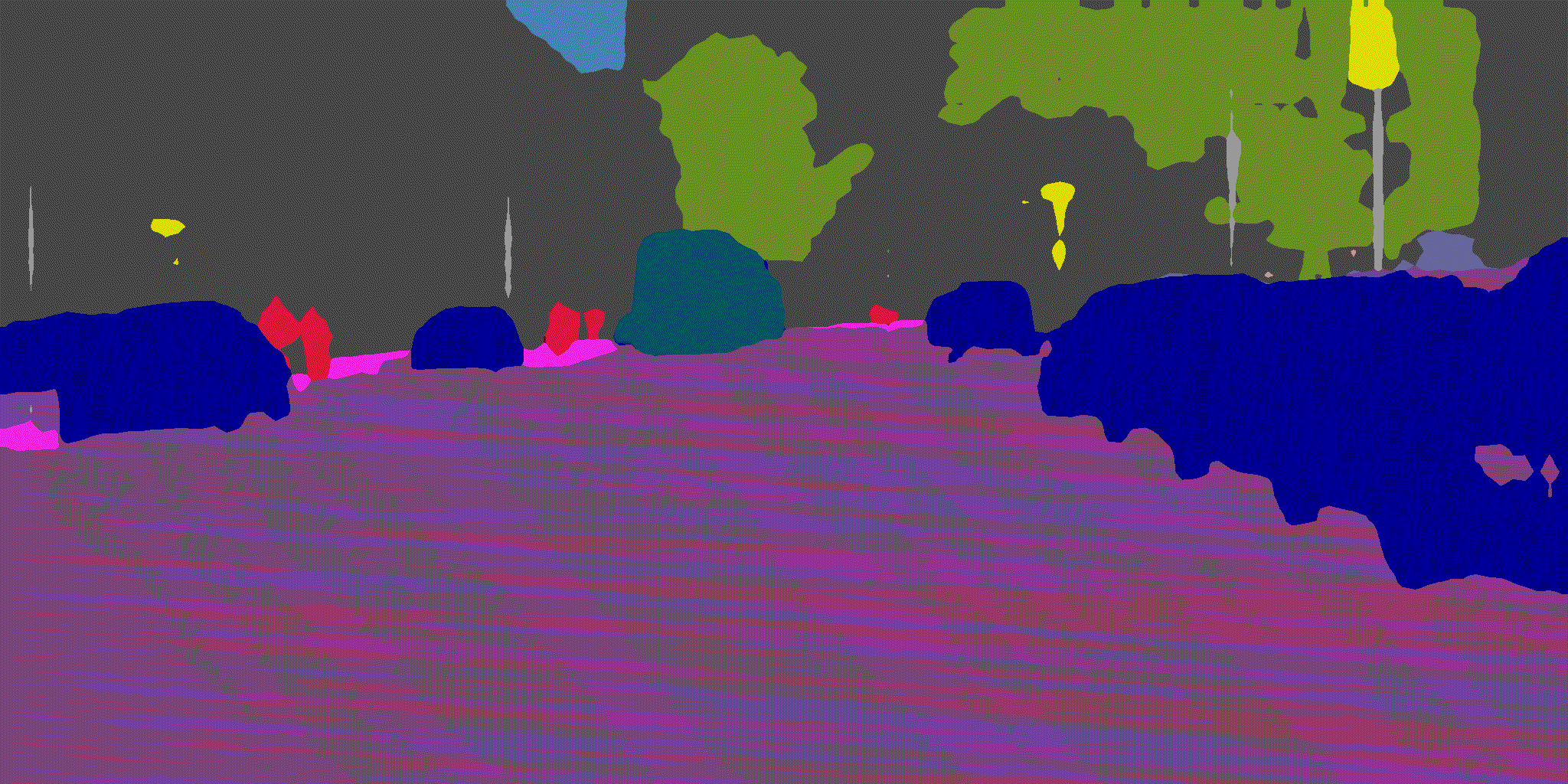}\\        
			
			\multirow{-5}{*}{} \quad & 
			\multicolumn{3}{c}{\includegraphics[width=0.9\linewidth]{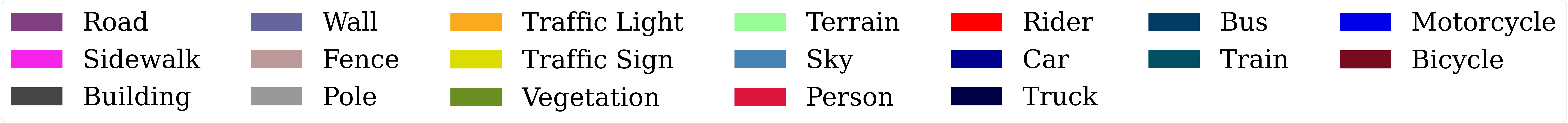}} \\
			
			\multirow{-5}{*}{\rotatebox[origin=c]{90}{\small GT}} \quad & 
			\includegraphics[width=.32\linewidth]{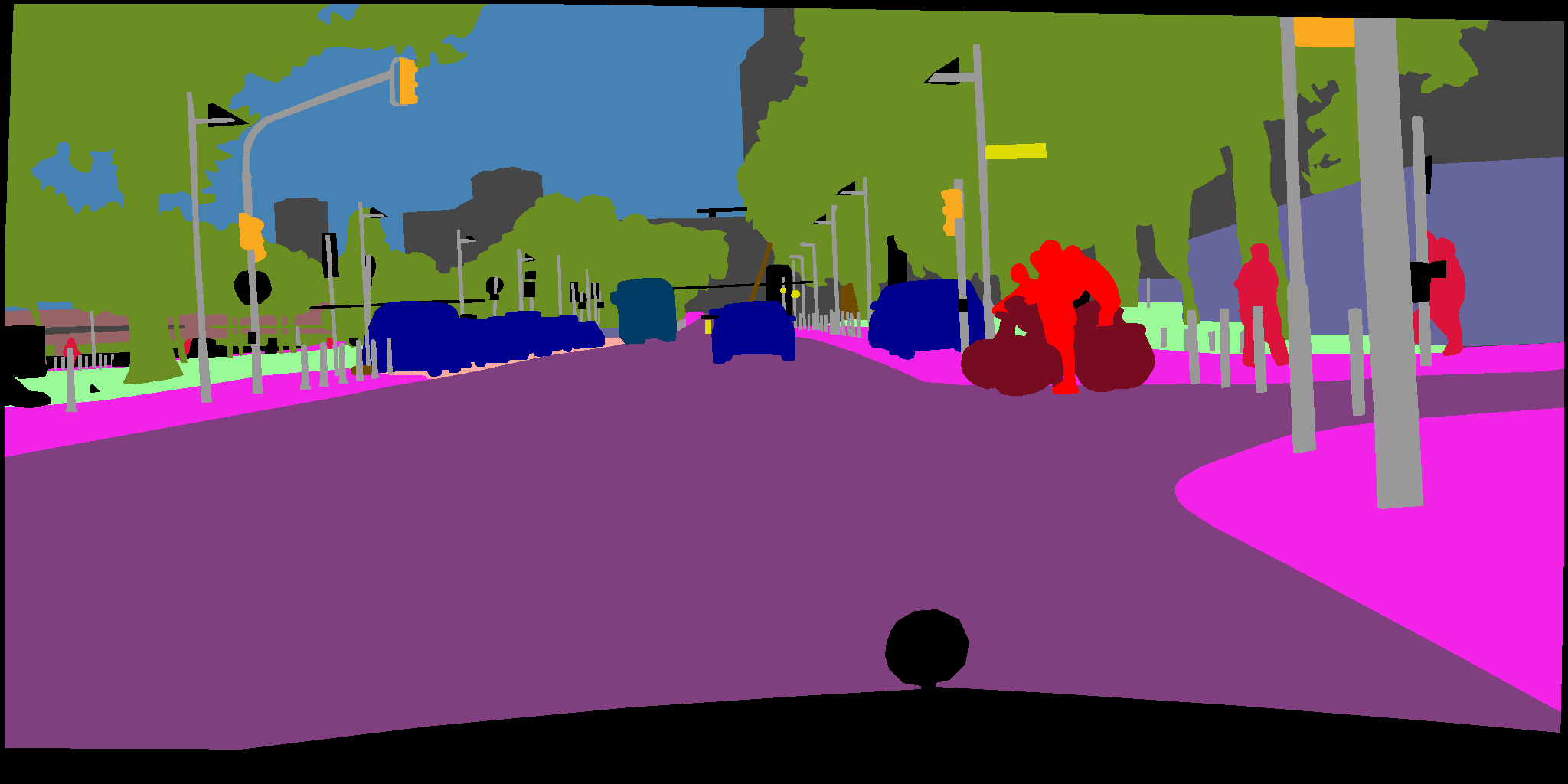} & \includegraphics[width=.32\linewidth]{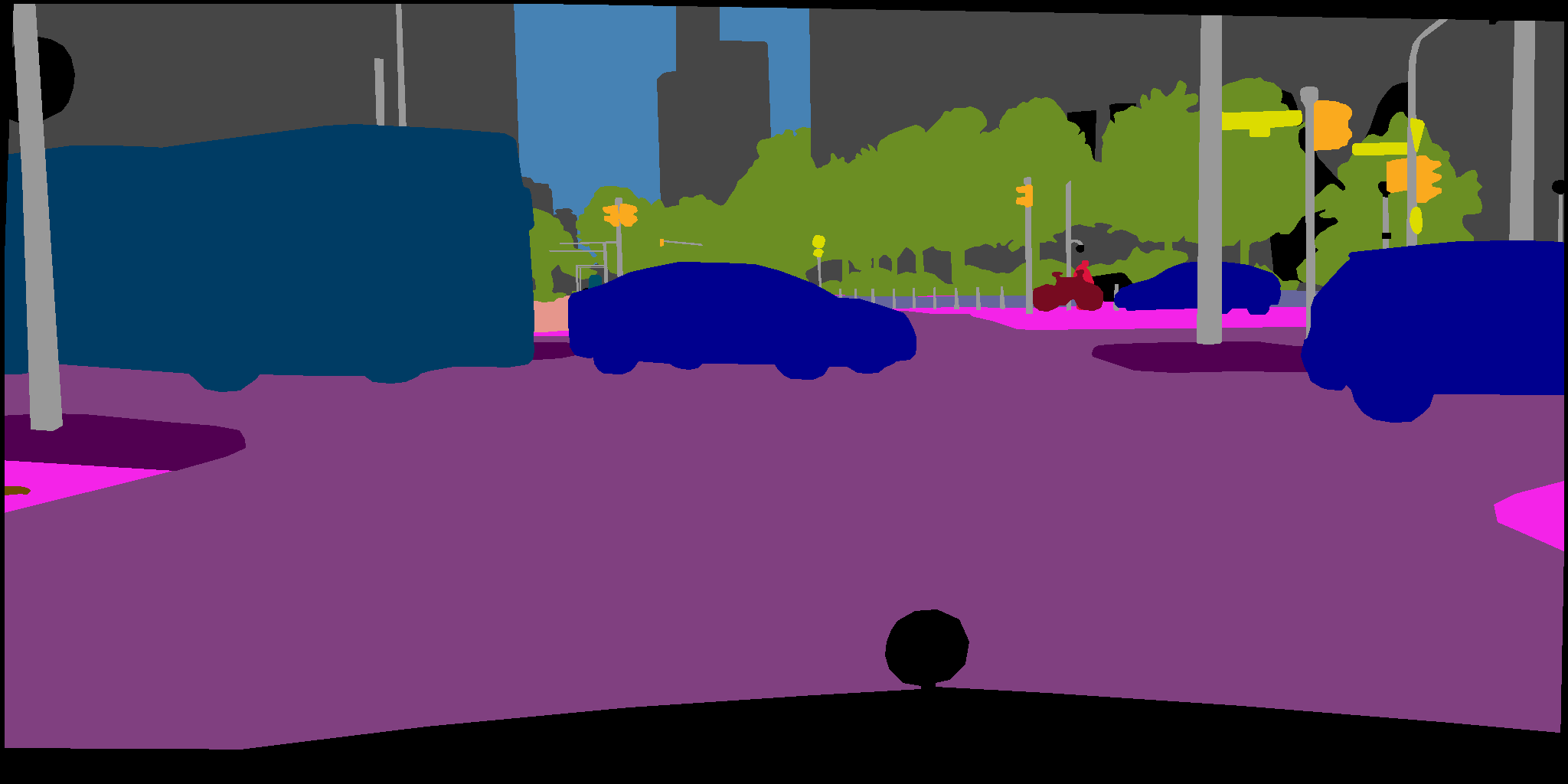}& \includegraphics[width=.32\linewidth]{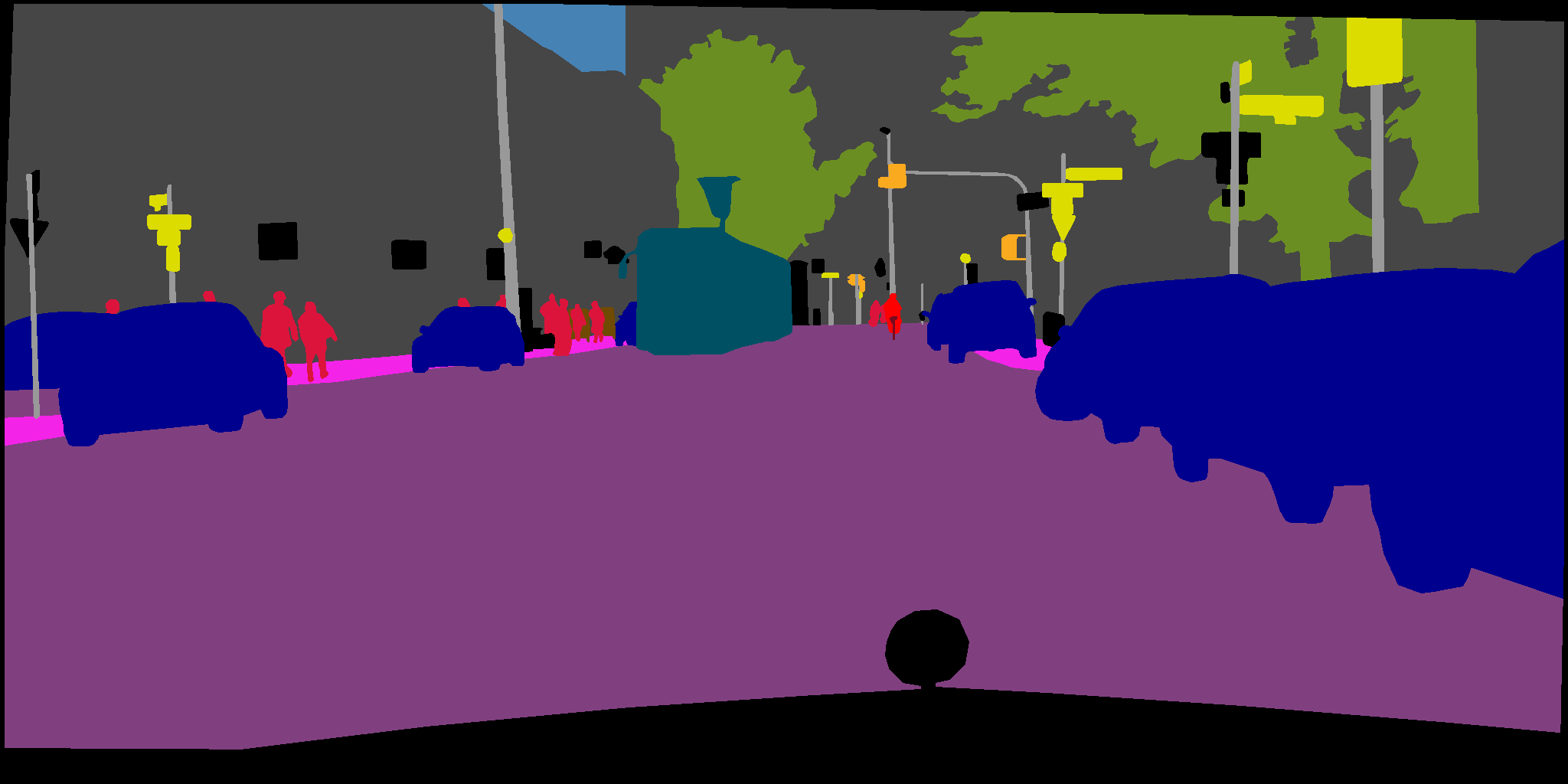}\\    
			
		\end{tabular}
	\end{adjustbox}
	\caption{Task-wise visualization for CLEO setting \textit{CS-Ex1} on the Cityscapes dataset \cite{cordts2016cityscapes}. The sequential refinement of classes can be observed at each evolutionary step.}
	\label{fig:cs_ex1_taskwise}
\end{figure}

\Cref{tab:cs_ex1_task_results} shows task-wise results on Cityscapes after training on the final task.
We have computed the mIoU for the classes of each task and compare our approach MoOn to MiB \cite{MiB}, PLOP \cite{PLOP}, and RCIL \cite{RCIL}.
Fine-tuning typically leads to catastrophic forgetting of previous tasks, and only information of classes learned in the final task is retained. 
However for \textit{CS-Ex1} not even this is the case, since the last task contains the highly under-represented class \textit{motorcycle}.
All CL approaches struggle with task 3 and 5 In this experiment for the same reason.
MoOn demonstrates the lowest amount of forgetting of the initial set of classes in task 0 and the highest plasticity when learning new classes throughout the evolution of the ontology.
\Cref{fig:cs_ex1_taskwise} visualizes the results of MoOn for three validation samples of Cityscapes to show how the semantic ontology evolves over time in \textit{CS-Ex1}.
The figure visualizes the 19 classes that are grouped into 7 parent classes in task 0.
In task 1, the road class is split from the parent class \textit{flat}, retaining the \textit{sidewalk} class, and the car class is split from the parent class \textit{vehicle}.
The \textit{vehicle} class continues to split in subsequent tasks, such as into the \textit{bus} class in task 3.

\begin{table}[t]
	\centering
	\caption{Task-wise results in mIoU for CS-Ex1 after learning the final task.}
	\label{tab:cs_ex1_task_results}
	\begin{tabular}{cccccccc}
		\boldhline
		\textbf{Method} & \textbf{Task 0} & \textbf{Task 1} & \textbf{Task 2} & \textbf{Task 3} & \textbf{Task 4} & \textbf{Task 5} & \textbf{All}\\ \boldhline
		Fine-Tuning & 00.00 & 00.00 & 00.00 & 00.00 & 00.00 & 00.00 & 00.00 \\ \hline
		Joint Training & 47.31 & 73.84 & 45.65 & 64.03 & 48.76 & 41.08 & 55.62 \\ \boldhline
		MiB \cite{MiB} & 17.27 & 03.76 & 02.35 & 00.00 & 14.32 & 00.00 & 09.11 \\ \hline
		PLOP \cite{PLOP} & 20.72 & 68.45 & 00.24 & 00.00 & 00.93 & 00.00 & 28.91 \\ \hline
		RCIL \cite{RCIL} & 17.20 & 00.00 & 00.00 & 00.00 & 00.00 & 00.00 & 06.88 \\ \hline
		MoOn (Ours) & 39.02 & 71.59 & 10.43 & 00.72 & 12.44 & 00.00 & \textbf{39.31}\\ \boldhline
	\end{tabular}
\end{table}

\section{Class-wise Results}

\Cref{tab:cs_ex1_classwise_results,tab:cs_ex2_classwise_results} show the class-wise results of our approach MoOn after each task.
The results shed light on the influence of the splitting order.
In case of \textit{CS-Ex1}, one class is split from each of the parent classes whereas in \textit{CS-Ex2}, each parent class is split completely in each step.
We can observe that in \textit{CS-Ex2}, the results for the \textit{Vehicle} subclasses are better compared to \textit{CS-Ex1}, as it splits completely in a single final step.
On the contrary, we observe that a few classes which are split in the earlier tasks achieve better results in \textit{CS-Ex1}, \eg the \textit{wall} or \textit{rider} classes.

\begin{table}
	\centering
	\caption{Class-wise results for CS-Ex1 after learning each task.}
	\label{tab:cs_ex1_classwise_results}
	\begin{adjustbox}{width=\textwidth}
	\begin{tabular}{c||c|c||c|c|c||c|c|c||c|c||c||c|c||c|c|c|c|c|c}
		\boldhline
         & \rotatebox{90}{\textit{road}} & \rotatebox{90}{\textit{sidewalk}} & \rotatebox{90}{\textit{building}} & \rotatebox{90}{\textit{wall}} & \rotatebox{90}{\textit{fence}} & \rotatebox{90}{\textit{pole}} & \rotatebox{90}{\textit{traffic light}} & \rotatebox{90}{\textit{traffic sign}} & \rotatebox{90}{\textit{vegetation}} & \rotatebox{90}{\textit{terrain}} & \rotatebox{90}{\textit{sky}} & \rotatebox{90}{\textit{person}} & \rotatebox{90}{\textit{rider}} & \rotatebox{90}{\textit{car}} & \rotatebox{90}{\textit{truck}} & \rotatebox{90}{\textit{bus}} & \rotatebox{90}{\textit{train}} & \rotatebox{90}{\textit{motorcycle}} & \rotatebox{90}{\textit{bicycle}} \\
         
         \hline
		
		\textbf{Task} & \multicolumn{2}{c||}{\textbf{Flat}} & \multicolumn{3}{c||}{\textbf{Construction}} & \multicolumn{3}{c||}{\textbf{Object}} & \multicolumn{2}{c||}{\textbf{Nature}} & \textbf{Sky} & \multicolumn{2}{c||}{\textbf{Human}} & \multicolumn{6}{c}{\textbf{Vehicle}} \\
        
        \boldhline
		
		0 & \multicolumn{2}{c||}{97.99} & \multicolumn{3}{c||}{86.78} & \multicolumn{3}{c||}{33.74} & \multicolumn{2}{c||}{88.58} & 90.40 & \multicolumn{2}{c||}{65.65} & \multicolumn{6}{c}{85.01} \\ \hline
		
		1 & \multicolumn{1}{c|}{96.95} & 59.11 & \multicolumn{1}{c|}{86.44} & \multicolumn{2}{c||}{36.94} & \multicolumn{1}{c|}{19.49} & \multicolumn{2}{c||}{42.29} & \multicolumn{1}{c|}{88.15} & 41.61 & 90.41 & \multicolumn{1}{c|}{61.65} & 15.23 & \multicolumn{1}{c|}{83.59} & \multicolumn{5}{c}{58.90} \\ \hline
		
		2 & \multicolumn{1}{c|}{96.93} & 59.18 & \multicolumn{1}{c|}{86.32} & \multicolumn{1}{c|}{31.92} & 22.47 & \multicolumn{1}{c|}{19.35} & \multicolumn{1}{c|}{00.00} & 39.19 & \multicolumn{1}{c|}{88.13} & 41.76 & 90.29 & \multicolumn{1}{c|}{61.47} & 14.89 & \multicolumn{1}{c|}{82.78} & \multicolumn{1}{c|}{23.69} & \multicolumn{4}{c}{50.90} \\ \hline
		
		3 & \multicolumn{1}{c|}{96.88} & 58.91 & \multicolumn{1}{c|}{86.26} & \multicolumn{1}{c|}{32.06} & 21.42 & \multicolumn{1}{c|}{17.90} & \multicolumn{1}{c|}{00.00} & 38.34 & \multicolumn{1}{c|}{88.09} & 41.81 & 90.21 & \multicolumn{1}{c|}{61.24} & 14.68 & \multicolumn{1}{c|}{83.27} & \multicolumn{1}{c|}{00.00} & \multicolumn{1}{c|}{37.73} & \multicolumn{3}{c}{44.38} \\ \hline
		
		4 & \multicolumn{1}{c|}{96.84} & 58.56 & \multicolumn{1}{c|}{86.19} & \multicolumn{1}{c|}{28.10} & 20.21 & \multicolumn{1}{c|}{16.15} & \multicolumn{1}{c|}{00.00} & 37.48 & \multicolumn{1}{c|}{87.84} & 41.69 & 90.17 & \multicolumn{1}{c|}{60.63} & 15.55 & \multicolumn{1}{c|}{82.10} & \multicolumn{1}{c|}{00.00} & \multicolumn{1}{c|}{1.42} & \multicolumn{1}{c|}{13.11} & \multicolumn{2}{c}{50.55} \\ \hline
		
		5 & \multicolumn{1}{c|}{96.79} & 58.17 & \multicolumn{1}{c|}{86.15} & \multicolumn{1}{c|}{31.31} & 21.72 & \multicolumn{1}{c|}{15.65} & \multicolumn{1}{c|}{00.00} & 37.92 & \multicolumn{1}{c|}{88.00} & 41.81 & 90.21 & \multicolumn{1}{c|}{60.39} & 13.01 & \multicolumn{1}{c|}{82.59} & \multicolumn{1}{c|}{00.00} & \multicolumn{1}{c|}{00.72} & \multicolumn{1}{c|}{12.44} & \multicolumn{1}{c|}{00.00} & 49.38 \\ \boldhline	
	
	\end{tabular}
\end{adjustbox}
\end{table}

\begin{table}
	\centering
	\caption{Class-wise results for CS-Ex2 after learning each task.}
	\label{tab:cs_ex2_classwise_results}
	\begin{adjustbox}{width=\textwidth}
	\begin{tabular}{c||c|c||c|c|c||c|c|c||c|c||c||c|c||c|c|c|c|c|c}
		\boldhline
         & \rotatebox{90}{\textit{road}} & \rotatebox{90}{\textit{sidewalk}} & \rotatebox{90}{\textit{building}} & \rotatebox{90}{\textit{wall}} & \rotatebox{90}{\textit{fence}} & \rotatebox{90}{\textit{pole}} & \rotatebox{90}{\textit{traffic light}} & \rotatebox{90}{\textit{traffic sign}} & \rotatebox{90}{\textit{vegetation}} & \rotatebox{90}{\textit{terrain}} & \rotatebox{90}{\textit{sky}} & \rotatebox{90}{\textit{person}} & \rotatebox{90}{\textit{rider}} & \rotatebox{90}{\textit{car}} & \rotatebox{90}{\textit{truck}} & \rotatebox{90}{\textit{bus}} & \rotatebox{90}{\textit{train}} & \rotatebox{90}{\textit{motorcycle}} & \rotatebox{90}{\textit{bicycle}} \\
         
         \hline
         
	\textbf{Task} & \multicolumn{2}{c||}{\textbf{Flat}} & \multicolumn{3}{c||}{\textbf{Construction}} & \multicolumn{3}{c||}{\textbf{Object}} & \multicolumn{2}{c||}{\textbf{Nature}} & \textbf{Sky} & \multicolumn{2}{c||}{\textbf{Human}} & \multicolumn{6}{c}{\textbf{Vehicle}} \\ 
        
        \boldhline
	
		0 & \multicolumn{2}{c||}{97.99} & \multicolumn{3}{c||}{86.78} & \multicolumn{3}{c||}{33.74} & \multicolumn{2}{c||}{88.58} & 90.40 & \multicolumn{2}{c||}{65.65} & \multicolumn{6}{c}{85.01} \\ \hline 
		
		1 & \multicolumn{1}{c|}{96.99} & 59.25 & \multicolumn{3}{c||}{86.76} & \multicolumn{3}{c||}{33.79} & \multicolumn{2}{c|}{88.56} & 90.39 & \multicolumn{2}{c||}{65.49} & \multicolumn{6}{c}{84.97} \\ \cline{2-20} 
	
		2 & \multicolumn{1}{c|}{96.98} & 59.05 & \multicolumn{1}{c|}{86.51} & \multicolumn{1}{c|}{24.21} & 21.91 & \multicolumn{3}{c||}{34.01} & \multicolumn{2}{c||}{88.54} & 90.41 & \multicolumn{2}{c||}{65.35} & \multicolumn{6}{c}{84.72} \\\hline
	
		3 & \multicolumn{1}{c|}{96.98} & 59.01 & \multicolumn{1}{c|}{86.38} & \multicolumn{1}{c|}{24.17} & 22.33 & \multicolumn{1}{c|}{20.23} & \multicolumn{1}{c|}{00.00} & 37.2 & \multicolumn{2}{c||}{88.42} & 90.29 & \multicolumn{2}{c||}{65.01} & \multicolumn{6}{c}{84.67} \\ \hline 
	
		4 & \multicolumn{1}{c|}{96.95} & 58.65 & \multicolumn{1}{c|}{86.38} & \multicolumn{1}{c|}{24.65} & 21.98 & \multicolumn{1}{c|}{19.38} & \multicolumn{1}{c|}{00.00} & 37.35 & \multicolumn{1}{c|}{88.18} & 41.05 & 90.32 & \multicolumn{2}{c||}{64.88} & \multicolumn{6}{c}{84.55} \\ \hline 
	
		5 & \multicolumn{1}{c|}{96.93} & 58.35 & \multicolumn{1}{c|}{86.29} & \multicolumn{1}{c|}{25.59} & 21.44 & \multicolumn{1}{c|}{18.73} & \multicolumn{1}{c|}{00.00} & 37.22 & \multicolumn{1}{c|}{88.18} & 40.88 & 90.29 & \multicolumn{1}{c|}{61.24} & 02.31 & \multicolumn{6}{c}{84.44} \\ \hline
	
		6 & \multicolumn{1}{c|}{96.89} & 58.36 & \multicolumn{1}{c|}{86.13} & \multicolumn{1}{c|}{26.40} & 22.12 & \multicolumn{1}{c|}{18.38} & \multicolumn{1}{c|}{00.00} & 37.52 & \multicolumn{1}{c|}{88.17} & 40.55 & 90.24 & \multicolumn{1}{c|}{60.97} & 06.15 & \multicolumn{1}{c|}{80.47} & \multicolumn{1}{c|}{01.64} & \multicolumn{1}{c|}{25.96} & \multicolumn{1}{c|}{22.44} & \multicolumn{1}{c|}{00.00} & 50.00 \\ \boldhline
	\end{tabular}
\end{adjustbox}
\end{table}

\section{Task-wise Semantic Evolution}
For maximum transparency, we explicitly list the classes $C_t$ for each task of each of our experiments.
For the two experimental setting on Cityscapes, the evolution of classes is described in \cref{tab:app:csex1,tab:app:csex2}, for PASCAL VOC, the evolution is given in \cref{tab:app:vocex1,tab:app:vocex2,tab:app:vocex3}, and for Mapillary Vistas, we list all evolutionary steps in \cref{tab:app:mvex1,tab:app:mvex2}.
Additionally, in each table, we list all the classes contained in each evaluation category (\textit{unsplit}, \textit{split}, \textit{retained}) after the sequential learning.

\begin{table}[t]
	\centering
	\caption{Evolutionary steps for CS-Ex1.}
	\label{tab:app:csex1}
	\begin{tabular}{p{0.125\linewidth} p{0.85\linewidth}}
		\boldhline
		\multicolumn{2}{l}{\textbf{CS-Ex1}}\\
		\boldhline
		Task $t$ & $C_t$ \\
		\boldhline
		0 & \it background, flat, construction, object, nature, sky, human, vehicle\\
		\hline
		1 & \it road, building, pole, vegetation, person, car\\
		\hline
		2 & \it wall, traffic light, truck\\
		\hline
		3 & \it bus\\
		\hline
		4 & \it train\\
		\hline
		5 & \it motorcycle\\
		\boldhline
		\multicolumn{2}{l}{\textbf{Class Groups}}\\
		\boldhline
		Unsplit & \it sky \\ \hline
		Split & \it road,  building, pole, vegetation, person, car, wall, traffic light, truck, bus, train, motorcycle\\ \hline
		Retained & \it flat (sidewalk), construction (fence), object (traffic sign), nature (terrain), human (rider), vehicle (bicycle)\\
		\boldhline
	\end{tabular}
\end{table}

\begin{table}[t]
	\centering
	\caption{Evolutionary steps for CS-Ex2.}
	\label{tab:app:csex2}
	\begin{tabular}{p{0.125\linewidth} p{0.85\linewidth}}
		\boldhline
		\multicolumn{2}{l}{\textbf{CS-Ex2}}\\
		\boldhline
		Task $t$ & $C_t$ \\
		\boldhline
		0 & \it background, flat, construction, object, nature, sky, human, vehicle\\
		\hline
		1 & \it road\\
		\hline
		2 & \it building, wall\\
		\hline
		3 & \it pole, traffic light\\
		\hline
		4 & \it vegetation\\
		\hline
		5 & \it person\\
		\hline
		6 & \it car, truck, bus, train, motorcycle\\
		\boldhline
		\multicolumn{2}{l}{\textbf{Class Groups}}\\
		\boldhline
		Unsplit & \it sky \\ \hline
		Split & \it road,  building, wall, pole, traffic light, vegetation, person, car, truck, bus, train, motorcycle\\ \hline
		Retained & \it flat (sidewalk), construction (fence), object (traffic sign), nature (terrain), human (rider), vehicle (bicycle)\\
		\boldhline
	\end{tabular}
\end{table}


\begin{table}[t]
	\centering
	\caption{Evolutionary steps for VOC-Ex1.}
	\label{tab:app:vocex1}
	\begin{tabular}{p{0.125\linewidth} p{0.85\linewidth}}
		\boldhline
		\multicolumn{2}{l}{\textbf{VOC-Ex1}}\\
		\boldhline
		Task $t$ & $C_t$ \\
		\boldhline
		0 & \it background, animals, household, person, vehicle\\
		\hline
		1 & \it farmyard, bird, bottle, furniture, 2-wheeler, aeroplane\\
		\hline
		2 & \it cow, horse, sheep, chair, sofa, dining table, bicycle, motorbike\\
		\boldhline
		\multicolumn{2}{l}{\textbf{Class Groups}}\\
		\boldhline
		Unsplit & \it person \\ \hline
		Split & \it bird, bottle, aeroplane, cow, horse, sheep, chair, sofa, dining table, bicycle, motorbike\\ \hline
		Retained & \it animals (cat, dog), household (tv/monitor, plant), 4-wheeler (car, bus), vehicle (boat, train)\\
		\boldhline
	\end{tabular}
\end{table}

\begin{table}[t]
	\centering
	\caption{Evolutionary steps for VOC-Ex2.}
	\label{tab:app:vocex2}
	\begin{tabular}{p{0.125\linewidth} p{0.85\linewidth}}
		\boldhline
		\multicolumn{2}{l}{\textbf{VOC-Ex2}}\\
		\boldhline
		Task $t$ & $C_t$ \\
		\boldhline
		0 & \it background, animals, household, person, vehicle\\
		\hline
		1 & \it bird, plant, train\\
		\hline
		2 & \it sheep, tv/monitor, boat\\
		\hline
		3 & \it horse, dining table, aeroplane\\
		\hline
		4 & \it cow, sofa, motorbike\\
		\hline
		5 & \it dog, chair, bicycle\\
		\boldhline
		\multicolumn{2}{l}{\textbf{Class Groups}}\\
		\boldhline
		Unsplit & \it person \\ \hline
		Split & \it bird, plant, train, sheep, tv/monitor, boat, horse, dining table, aeroplane, cow, sofa, motorbike, dog, chair, bicycle\\ \hline
		Retained & \it animals (cat), household (bottle), vehicle (car, bus)\\
		\boldhline
	\end{tabular}
\end{table}

\begin{table}[t]
	\centering
	\caption{Evolutionary steps for VOC-Ex3.}
	\label{tab:app:vocex3}
	\begin{tabular}{p{0.125\linewidth} p{0.85\linewidth}}
		\boldhline
		\multicolumn{2}{l}{\textbf{VOC-Ex3}}\\
		\boldhline
		Task $t$ & $C_t$ \\
		\boldhline
		0 & \it background, animals, household, person, vehicle\\
		\hline
		1 & \it dog, horse, cow, sheep, bird\\
		\hline
		2 & \it chair, sofa, table, tv/monitor, plant\\
		\hline
		3 & \it bus, bicycle, motorbike, aeroplane, boat, train\\
		\boldhline
		\multicolumn{2}{l}{\textbf{Class Groups}}\\
		\boldhline
		Unsplit & \it person \\ \hline
		Split & \it dog, horse, cow, sheep, bird, chair, sofa, table, tv/monitor, plant, bus, bicycle, motorbike, aeroplane, boat, train \\ \hline
		Retained & \it animals (cat), household (bottle), vehicle (car)\\
		\boldhline
	\end{tabular}
\end{table}

 \clearpage
\begin{longtable}{p{0.125\linewidth} p{0.85\linewidth}}
	\caption{Evolutionary steps for MV-Ex1.} \\
	\label{tab:app:mvex1}\\
	\boldhline
	\multicolumn{2}{l}{\textbf{MV-Ex1}}\\
	\boldhline
	Task $t$ & $C_t$ \\
	\boldhline
	0 & \it background, bird, ground animal, curb, fence, guard rail, barrier, wall, bike lane, crosswalk - plain, curb cut, parking, pedestrian area, rail track, road, service lane, sidewalk, bridge, building, tunnel, person, bicyclist, motorcyclist, other rider, lane marking - crosswalk, lane marking - general, mountain, sand, sky, snow, terrain, vegetation, water, banner, bench, bike rack, billboard, catch basin, cctv camera, fire hydrant, junction box, mailbox, manhole, phone booth, pothole, street light, pole, traffic sign frame, utility pole, traffic light, traffic sign (back), traffic sign (front), trash can, bicycle, boat, bus, car, caravan, motorcycle, on rails, other vehicle, trailer, truck, wheeled slow, car mount, ego vehicle\\
	\hline
	1 & \it ambiguous barrier, concrete block, driveway, dynamic, garage, ground, lane marking (only) - crosswalk, lane marking (only) - dashed line, lane marking (only) - other, lane marking (only) - test, lane marking - ambiguous, lane marking - arrow (left), lane marking - arrow (other), lane marking - arrow (right), lane marking - arrow (split left or straight), lane marking - arrow (split right or straight), lane marking - arrow (straight), lane marking - give way (row), lane marking - give way (single), lane marking - hatched (chevron), lane marking - hatched (diagonal), lane marking - other, lane marking - stop line, lane marking - straight line, lane marking - symbol (bicycle), lane marking - symbol (other), lane marking - text, lane marking - zigzag line, lane separator, parking aisle, parking meter, person group, pole group, road median, road shoulder, road side,  signage - ambiguous, signage - back, signage - information, signage - other, signage - store, static, temporary barrier, traffic cone, traffic island, traffic light - cyclists, traffic light - general (horizontal), traffic light - general (upright), traffic light - other, traffic light - pedestrians, traffic sign - ambiguous, traffic sign - direction (back), traffic sign - direction (front), traffic sign - parking, traffic sign - temporary (back), traffic sign - temporary (front), vehicle group, water valve
	\\
	\pagebreak
	\boldhline 
	\multicolumn{2}{l}{\textbf{Class Groups}}\\
	\boldhline
	
	Unsplit & \it bird, ground animal, curb, fence, guard rail,  wall, bike lane, crosswalk - plain, curb cut, pedestrian area, rail track, service lane, sidewalk, bridge, building, tunnel, person, bicyclist, motorcyclist, other rider, mountain, sand, sky, snow, terrain, vegetation, water, banner, bench, bike rack,  catch basin, cctv camera, fire hydrant, junction box, mailbox, manhole, phone booth, pothole, street light, pole, traffic sign frame, utility pole, trash can, bicycle, boat, bus, car, caravan, motorcycle, on rails, other vehicle, trailer, truck, wheeled slow, car mount, ego vehicle \\ \hline
	
	Split & \it ambiguous barrier, concrete block, driveway, dynamic, garage, ground, lane marking (only) - crosswalk, lane marking (only) - dashed line, lane marking (only) - other, lane marking (only) - test, lane marking - ambiguous, lane marking - arrow (left), lane marking - arrow (other), lane marking - arrow (right), lane marking - arrow (split left or straight), lane marking - arrow (split right or straight), lane marking - arrow (straight), lane marking - give way (row), lane marking - give way (single), lane marking - hatched (chevron), lane marking - hatched (diagonal), lane marking - other, lane marking - stop line, lane marking - straight line, lane marking - symbol (bicycle), lane marking - symbol (other), lane marking - text, lane marking - zigzag line, lane separator, parking aisle, parking meter, person group, pole group, road median, road shoulder, road side,  signage - ambiguous, signage - back, signage - information, signage - other, signage - store, static, temporary barrier, traffic cone, traffic island, traffic light - cyclists, traffic light - general (horizontal), traffic light - general (upright), traffic light - other, traffic light - pedestrians, traffic sign - ambiguous, traffic sign - direction (back), traffic sign - direction (front), traffic sign - parking, traffic sign - temporary (back), traffic sign - temporary (front), vehicle group, water valve \\ \hline
	
	Retained & \it background, barrier, lane marking - crosswalk, parking, road, traffic sign (front), traffic sign (back), signage - advertisement, lane marking - dashed line, traffic light - general (single) \\
	\boldhline
	
\end{longtable}

\clearpage
\begin{longtable}{p{0.125\linewidth} p{0.85\linewidth}}
	\caption{Evolutionary steps for MV-Ex2.} \\
	\label{tab:app:mvex2}\\
	\boldhline
	\multicolumn{2}{l}{\textbf{MV-Ex2}}\\
	\boldhline
	Task $t$ & $C_t$ \\
	\boldhline
	0 & \it background, bird, ground animal, curb, fence, guard rail, barrier, wall, bike lane, crosswalk - plain, curb cut, parking, pedestrian area, rail track, road, service lane, sidewalk, bridge, building, tunnel, person, bicyclist, motorcyclist, other rider, lane marking - crosswalk, lane marking - general, mountain, sand, sky, snow, terrain, vegetation, water, banner, bench, bike rack, billboard, catch basin, cctv camera, fire hydrant, junction box, mailbox, manhole, phone booth, pothole, street light, pole, traffic sign frame, utility pole, traffic light, traffic sign (back), traffic sign (front), trash can, bicycle, boat, bus, car, caravan, motorcycle, on rails, other vehicle, trailer, truck, wheeled slow, car mount, ego vehicle\\
	\hline
	1 & \it concrete block, road median, road side, lane separator \\
	\hline
	2 & \it lane marking (only) - crosswalk\\
	\hline
	3 & \it parking aisle\\
	\hline
	4 & \it road shoulder \\
	\hline
	5 & \it temporary barrier, traffic sign - direction (front), traffic sign - parking, traffic sign - temporary (front) \\
	\hline
	6 & \it traffic sign - direction (back), traffic sign - temporary (back) \\
	\hline
	7 & \it ambiguous barrier, driveway, traffic island, garage, person group, parking meter, pole group, traffic cone, traffic sign - ambiguous, vehicle group, water valve, dynamic, ground, static \\
	\hline
	8 & \it signage - ambiguous, signage - back, signage - information, signage - other, signage - store \\
	\hline
	9 & \it lane marking - straight line, lane marking - zigzag line, lane marking - ambiguous, lane marking - arrow (left), lane marking - arrow (other), lane marking - arrow (right), lane marking - arrow (split left or straight), lane marking - arrow (split right or straight), lane marking - arrow (straight), lane marking - give way (row), lane marking - give way (single), lane marking - hatched (chevron), lane marking - hatched (diagonal), lane marking - other, lane marking - stop line, lane marking - symbol (bicycle), lane marking - symbol (other), lane marking - text, lane marking (only) - dashed line, lane marking (only) - other, lane marking (only) - test\\
	\hline
	10 & \it traffic light - pedestrians, traffic light - general (upright), traffic light - general (horizontal), traffic light - cyclists, traffic light - other\\
	\pagebreak
	\boldhline
	\multicolumn{2}{l}{\textbf{Class Groups}}\\
	\boldhline
	
	Unsplit & \it bird, ground animal, curb, fence, guard rail,  wall, bike lane, crosswalk - plain, curb cut, pedestrian area, rail track, service lane, sidewalk, bridge, building, tunnel, person, bicyclist, motorcyclist, other rider, mountain, sand, sky, snow, terrain, vegetation, water, banner, bench, bike rack,  catch basin, cctv camera, fire hydrant, junction box, mailbox, manhole, phone booth, pothole, street light, pole, traffic sign frame, utility pole, trash can, bicycle, boat, bus, car, caravan, motorcycle, on rails, other vehicle, trailer, truck, wheeled slow, car mount, ego vehicle \\
	\hline
	
	Split & \it ambiguous barrier, concrete block, driveway, dynamic, garage, ground, lane marking (only) - crosswalk, lane marking (only) - dashed line, lane marking (only) - other, lane marking (only) - test, lane marking - ambiguous, lane marking - arrow (left), lane marking - arrow (other), lane marking - arrow (right), lane marking - arrow (split left or straight), lane marking - arrow (split right or straight), lane marking - arrow (straight), lane marking - give way (row), lane marking - give way (single), lane marking - hatched (chevron), lane marking - hatched (diagonal), lane marking - other, lane marking - stop line, lane marking - straight line, lane marking - symbol (bicycle), lane marking - symbol (other), lane marking - text, lane marking - zigzag line, lane separator, parking aisle, parking meter, person group, pole group, road median, road shoulder, road side,  signage - ambiguous, signage - back, signage - information, signage - other, signage - store, static, temporary barrier, traffic cone, traffic island, traffic light - cyclists, traffic light - general (horizontal), traffic light - general (upright), traffic light - other, traffic light - pedestrians, traffic sign - ambiguous, traffic sign - direction (back), traffic sign - direction (front), traffic sign - parking, traffic sign - temporary (back), traffic sign - temporary (front), vehicle group, water valve \\
	\hline
	
	Retained & \it background, barrier, lane marking - crosswalk, parking, road, traffic sign (front), traffic sign (back), signage - advertisement, lane marking - dashed line, traffic light - general (single) \\
	\boldhline	
\end{longtable}

\end{appendix}

%
%
\bibliographystyle{splncs04}
\bibliography{main}
\end{document}